\documentclass[11pt,letterpaper]{article}
\usepackage{liveaigc_report}
\usepackage[utf8]{inputenc}
\usepackage{xspace}
\usepackage[greek,english]{babel}
\definecolor{casegray}{HTML}{7B7B7B}
\definecolor{casepurple}{HTML}{7D6985}
\definecolor{casebrown}{HTML}{AA896B}
\definecolor{caseteal}{HTML}{006C70}
\newtcolorbox{agentcase}[4][]{enhanced,breakable,width=\dimexpr\linewidth-#3\relax,
  left skip=#3,colback=black!1,colframe=#2,colbacktitle=#2,coltitle=white,
  title={#4},title after break={#4 (continued)},fonttitle=\sffamily\bfseries\fontsize{9}{10}\selectfont,
  fontupper=\fontsize{9.3}{10.6}\selectfont\raggedright\setlength{\parindent}{0pt},arc=6pt,boxrule=0.4pt,
  left=7pt,right=7pt,top=3pt,bottom=3pt,toptitle=2pt,bottomtitle=2pt,
  before skip=2pt,after skip=2pt,parbox=false,#1}
\newcommand{\trajectoryend}{\par\smallskip\noindent\hfill
  \tcbox[on line,colback=black!2,colframe=casegray,boxrule=0.4pt,arc=3pt,
    left=5pt,right=5pt,top=2pt,bottom=2pt]{\sffamily\small End of trajectory.}\par}
\newtcolorbox{promptcase}[2]{enhanced,breakable,colback=black!1,
  colframe=#1,colbacktitle=#1,coltitle=white,title={#2},title after break={#2 (continued)},
  fonttitle=\sffamily\bfseries\small,fontupper=\fontsize{10}{11.5}\selectfont\raggedright\setlength{\parindent}{0pt},
  arc=6pt,boxrule=0.4pt,left=8pt,right=8pt,top=6pt,bottom=6pt,
  before skip=7pt,after skip=7pt,parbox=false}
\newcommand{\method}{\mbox{\textsc{MultiAgent-OPD}}\xspace}
\newcommand{\leader}{\textsc{Leader}\xspace}
\newcommand{\searcher}{\textsc{Searcher}\xspace}
\newcommand{\searchers}{\textsc{Searchers}\xspace}
\newcommand{\answerer}{\textsc{Answerer}\xspace}
\newcommand{\teacher}{\textsc{Teacher}\xspace}
\newif\ifshowedits
\showeditsfalse
\newcommand{\edit}[1]{\ifshowedits{\color{blue}#1}\else{\color{black}#1}\fi}

\pretocmd{\subsection}{\FloatBarrier\Needspace{5\baselineskip}}{}{}
\title{MultiAgent-OPD: End-to-End Training of Hierarchical Multi-Agent Systems with On-Policy Distillation}
\shorttitle{MultiAgent-OPD}
\reportlabel{Technical Report}
\reportdate{October 2026}
\leftlogo[height=0.9cm]{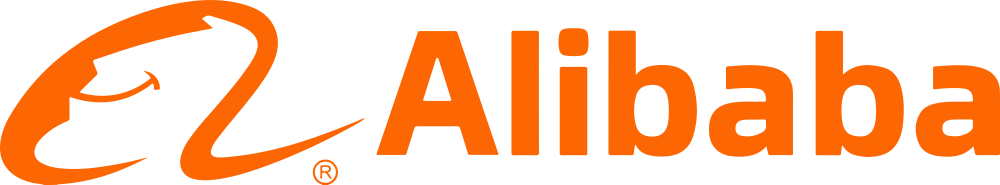}
\rightlogo[height=0.9cm,trim=436bp 638bp 306bp 480bp,clip]{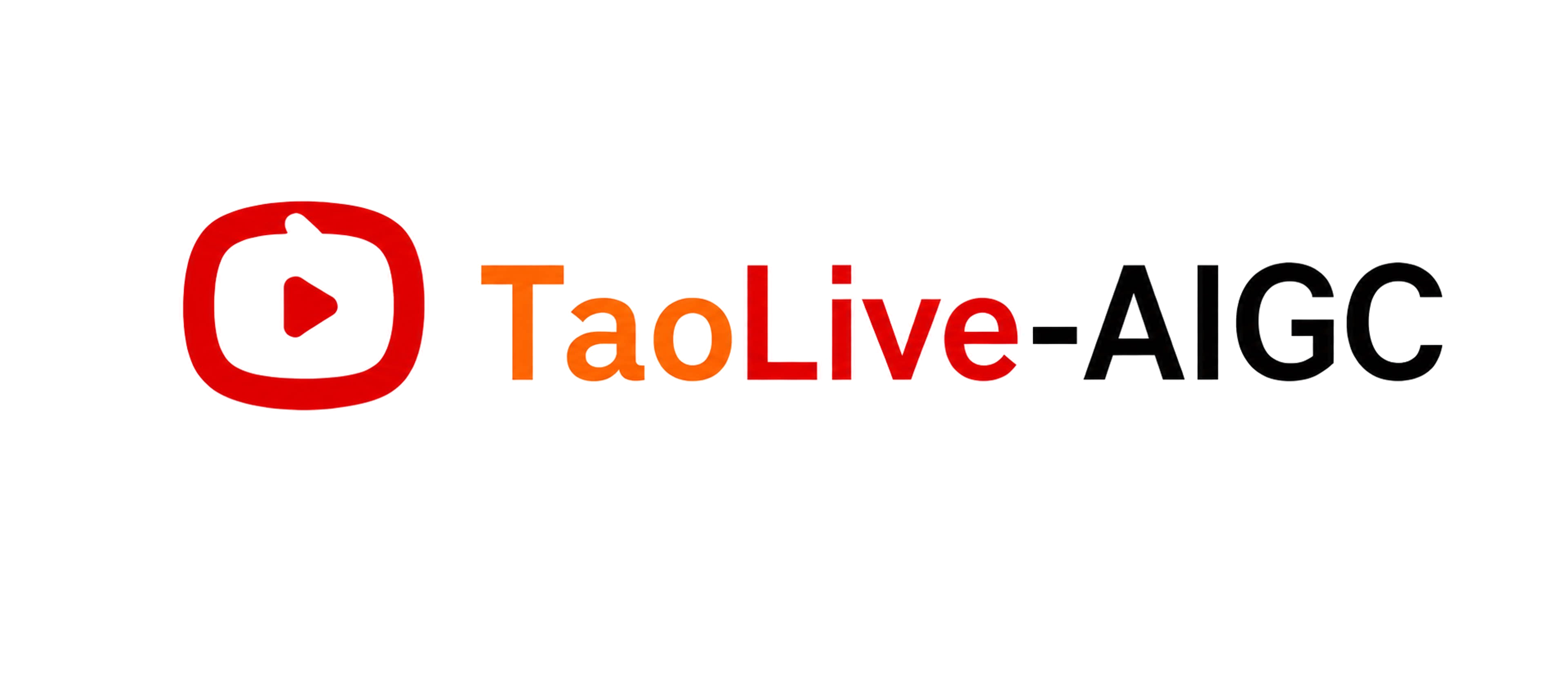}
\author{Yuhan Sun$^{1,\ddagger}$, Jiaxin Zhao$^{1,2,\dagger,\ddagger}$, Yongdong Luo$^{1}$, Yibo Hu$^{1,*}$,\\[3pt]
Meiguang Jin$^{1,*}$, Junfeng Ma$^{1}$, Weixun Wang$^{1}$, Weihang Pan$^{2}$, Zhongan Bi$^{2}$}

\hypersetup{pdftitle={MultiAgent-OPD: End-to-End Training of Hierarchical Multi-Agent Systems with On-Policy Distillation},
pdfauthor={Yuhan Sun, Jiaxin Zhao, Yongdong Luo, Yibo Hu, Meiguang Jin, Junfeng Ma, Weixun Wang, Weihang Pan, Zhongan Bi},
pdfsubject={Hierarchical multi-agent training with on-policy distillation},
pdfkeywords={Multi-agent systems, Reinforcement learning, On-policy distillation}}

\begin{document}
\maketitle
\begingroup
\renewcommand{\thefootnote}{\fnsymbol{footnote}}
\footnotetext[2]{Work done during an internship at Alibaba Group.}
\footnotetext[3]{Equal contribution.}
\endgroup
\begin{abstract}
In a hierarchical multi-agent system, a \leader assigns tasks and subagents carry them out. Jointly training the \leader and the subagents is difficult because a final answer gives only indirect feedback to the subagents. An LLM judge can score the performance of the subagents, but these scores depend on the judge's own assessment of the quality of subagents' responses. We present \method, which combines on-policy distillation with reinforcement learning to jointly train a shared-parameter \leader--\searcher--\answerer team end to end. We first train a \teacher in a \searcher--\answerer workflow using answer correctness. During team training, the frozen \teacher scores student-generated search and summary tokens to provide local process supervision for the \searchers, while the \leader learns from team success and the \answerer from candidate correctness. Across seven datasets, the trained system reaches 38.68\% macro avg@4, exceeding Prompted Critic by 4.06 percentage points and C-MAS + RL, the trained cascaded multiagent system, by 2.61 percentage points. The team also solves more questions within four attempts.
\end{abstract}

\section{Introduction}
\label{sec:intro}

Reinforcement learning (RL) for LLM-based multi-agent systems faces a central
challenge: the final outcome reflects the team's combined decisions, making
it difficult to assess individual intermediate actions. As agents specialize in planning, information
acquisition, and reasoning, effective training requires signals that reflect
their different roles in a shared objective.
Recent advances in collaborative post-training span discussion-based RL
\citep{park2025maporl}, credit propagation through sequential reasoning
\citep{motwani2025malt}, and agent-wise advantage normalization
\citep{feng2026drmas}.

This difficulty is particularly pronounced in Leader--Subagent systems,
where a Leader decomposes a task, delegates its execution, and integrates
subagent reports. Kimi K2.5's PARL addresses the challenge by optimizing
the Leader with frozen subagents \citep{kimi2026k25}, \edit{so subagent adaptation stays} outside joint training. M-GRPO enables joint optimization by
combining final-answer and format rewards with LLM-based evaluations of
subtask execution \citep{hong2025mgrpo}. Such evaluations provide local
supervision, but their reliability depends on the judge's assessment
capabilities, which are subject to biases and reasoning errors
\citep{zheng2023judge}. Moreover, a single subtask score offers limited
guidance on the sequence of search decisions that produced the report.
Jointly optimizing the hierarchy calls for informative feedback
at the level of subagent execution.

We match learning signals to each role's relation to the final task objective. Outcome rewards align the \leader's coordination decisions with team success and directly evaluate the \answerer's predictions. \searchers, in contrast, acquire intermediate evidence whose downstream value depends on later selection and reasoning. A useful report may receive no positive outcome reward when another role fails to use it. This motivates us to introduce a complementary source of supervision for \searchers: a policy trained to acquire and summarize evidence guides them locally, while outcome rewards train coordination and answer generation.

\begin{figure*}[!htbp]
\centering
\includegraphics[width=\textwidth]{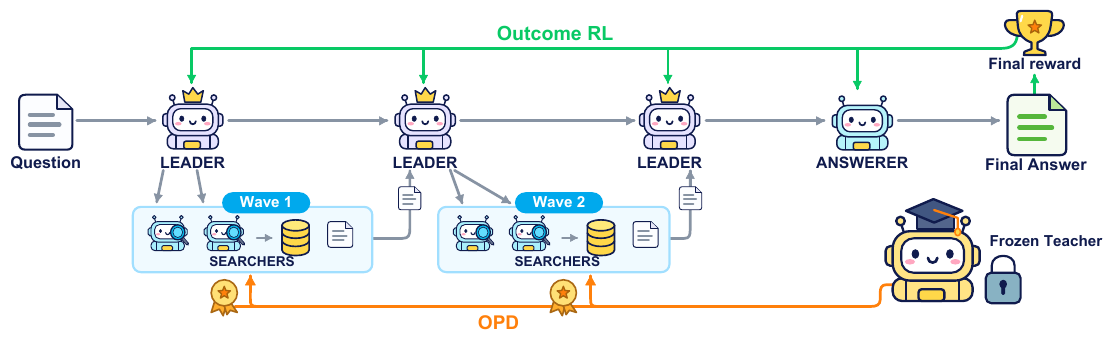}
\caption{\textbf{MultiAgent-OPD: multi-wave collaboration with two training signals.}
The same \leader assigns successive waves of \searchers, collects their
reports, and selects evidence for the \answerer. Each \searcher retrieves
evidence for its assigned subtask and summarizes the retrieved passages
into a report. Gray arrows show this
workflow. Orange OPD feedback from the frozen \teacher trains the \searchers\edit{, and} green final-answer reward feedback trains the \leader and the \answerer. All student
roles share one model. Two waves are shown for illustration.}
\label{fig:overview}
\end{figure*}

We introduce \method, a two-stage framework that integrates on-policy
distillation (OPD) into end-to-end RL for hierarchical multi-agent systems
(Figure~\ref{fig:overview}). The first stage trains the \teacher in a
simpler \searcher--\answerer workflow \edit{with rewards on answer correctness}. In the second stage, the frozen \teacher
provides token-level supervision on \searcher trajectories sampled from the
hierarchical student's current policy. Teacher--student log-probability
differences yield local advantages for query generation and evidence
summarization \citep{agarwal2024gkd,gu2024minillm}, while the \leader and the \answerer{} \edit{keep} outcome-based RL objectives. \edit{Because the \teacher conditions on the subtask and the interaction history of the student, the feedback follows the states that the team reaches as its policies evolve.} All three roles jointly update a shared student model\edit{, and} inference requires only the trained student.

On seven question-answering datasets, \method achieves 38.68\% macro avg@4, the highest among the evaluated systems. The score exceeds Prompted Critic (Appendix~\ref{sec:critic-prompt}) by 4.06 percentage points and C-MAS + RL, the cascaded multi-agent baseline, by 2.61 percentage points. It also exceeds the five RL-based search agents included in the comparison. Diagnostic comparisons examine the roles of direct \searcher supervision and task-specific \teacher training. Together, these results characterize how a policy learned in a simpler workflow can supply process supervision for subagents in a jointly trained hierarchy.

We make three contributions.
\begin{itemize}
\item We propose \method, a framework for end-to-end hierarchical multi-agent training that combines local \searcher supervision with outcome-based optimization of the \leader and the \answerer.
\item We formulate \searcher process supervision as on-policy distillation from a task-trained \teacher, which provides token-level guidance on the student's evolving subtask trajectories.
\item We report higher macro avg@4 across seven datasets than the included cascaded, hierarchical, and RL-based search baselines, and provide diagnostic comparisons of the \searcher learning signal, \teacher training, and direct subagent supervision.
\end{itemize}

\section{Related Work}

\paragraph{Multi-agent learning signals and credit assignment.}
MAPoRL trains agents through discussion, ReMA separates planning from
reasoning execution, and MHGPO uses heterogeneous rollout groups for
multi-agent search \citep{park2025maporl,wan2025rema,chen2026mhgpo}.
Dr.\ MAS uses agent-wise advantage normalization and verifier-controlled
Search--Answer calls \citep{feng2026drmas}. PARL trains a Leader with frozen
subagents \citep{kimi2026k25}. M-GRPO jointly trains main and subagent
models with outcome, format, and expert LLM scores \citep{hong2025mgrpo}.
Our method jointly trains the hierarchy with OPD from a task-trained
\teacher for \searcher supervision and outcome-based RL for the
\leader and \answerer.

\paragraph{Multi-agent systems.}
Agent harnesses such as Claude Code, Codex, DeepSeek Harness, and Kimi Code
support task delegation to subagents and integration of their results
\citep{anthropic2026subagents,openai2026subagents,deepseek2026harness,moonshot2026kimicode}.
Anthropic's research system similarly coordinates parallel searchers
\citep{anthropic2025research}. These systems motivate our Leader--Subagent
structure, and we study how to train the cooperating roles end to end.

\paragraph{On-policy distillation.}
GKD and MiniLLM train on student-generated sequences using teacher
probabilities \citep{agarwal2024gkd,gu2024minillm}.
Recent work shows how dense OPD feedback supports reasoning and preserves
learned behavior \citep{lu2025opd}. SOD adjusts distillation strength at
each tool-use step based on teacher--student divergence
\citep{zhong2026sod}, while MAD-OPD uses multi-agent debate to improve
teacher supervision \citep{wang2026madopd}. OPID derives hindsight skills from completed on-policy trajectories and uses them for self-distillation alongside outcome optimization \citep{yang2026opid}. SD-Search adds step-level supervision at query positions through hindsight self-distillation, where a single model acts as both student and outcome-conditioned teacher \citep{ma2026sdsearch}. OPDSearch+ distills a frozen off-the-shelf teacher on live search trajectories and then refines the student with RL \citep{ye2026opdsearchplus}. Both train a single search agent without task delegation. In contrast, we use a frozen, task-trained \searcher--\answerer policy to guide \searcher tokens on delegated subtask histories, while the \leader learns from team success and the \answerer from candidate correctness. Applying a task-trained teacher to delegated subtasks within a jointly trained hierarchy distinguishes our approach from prior work on OPD for search.

\section{MultiAgent-OPD}
\label{sec:method}

\edit{\method trains the \leader--\searcher--\answerer team in two stages and gives each
role a learning signal that matches the relation of the role to the final
outcome (Section~\ref{sec:intro}, Figure~\ref{fig:overview}). Stage~I trains
the simpler \searcher--\answerer workflow, in which the report goes straight to the
\answerer, and freezes the trained policy as the \teacher
(Section~\ref{sec:stage1}). Stage~II trains the team with role-specific
advantages, in which the \searchers learn from the \teacher through
on-policy distillation and the \leader and the \answerer learn from team
rewards (Section~\ref{sec:stage2}).}

\subsection{Preliminaries}
\label{sec:prelim}

\edit{The \leader breaks a question into subtasks, assigns the subtasks to
\searchers in successive waves and selects reports for the \answerer.
Each \searcher writes a query, reads the passages that the retriever
returns and then either searches again or writes a report. The \leader can
request further searches before the \answerer produces the final response,
so the subtask of each \searcher depends on the current policy of the
\leader. All roles share one language model $\pi_\theta$, and a role prompt
selects the role for each call. An update from any role also changes the policies of the other roles.}

\edit{For a question $x$ with reference answers $y^\star$, we sample a group
of $G$ trajectories $\tau_1,\dots,\tau_G$ with the sampling policy
$\pi_{\mathrm{old}}$, where a trajectory is one full run of a workflow on
$x$. We write $a_t$ for a generated token and $h_t$ for the history of the
token, which holds the role prompt, the question or subtask and every token
and retrieved passage before $a_t$ in the same call. We index the roles of
the \leader, the \searchers and the \answerer by $r\in\{L,S,A\}$ and
write $R_{r,i}$ for the reward of role $r$ in $\tau_i$.
During training, each answering request produces five candidates in
Stage~I, Stage~II, and all ablations. Let
$c_{i,j}$ be the normalized exact-match score of candidate $j$ in $\tau_i$
(Appendix~\ref{sec:appendix}). The team outcome is
$R_{\mathrm{EM},i}=\max_j c_{i,j}$: it equals 1 if at least one answer
candidate matches a reference answer under normalized exact match, and 0
otherwise. Each \answerer candidate receives its own exact-match reward. Training aims to raise the expected outcome
$\mathbb E_{x}\mathbb E_{\tau\sim\pi_\theta}[R_{\mathrm{EM}}]$ of the team,
and both stages start from the instruction-tuned base model $\pi_0$.}

\edit{The team outcome defines the task objective, while role-specific
rewards provide training signals for individual roles. Stage~I derives
advantages from these rewards. Stage~II retains reward-based advantages
for the \leader and \answerer and uses teacher-derived token-level
advantages for \searchers.}

\edit{We optimize the model with group-relative policy optimization (GRPO)
\citep{shao2024deepseekmath} and normalize the rewards within each role
\citep{feng2026drmas}. For workflow-level rewards, the advantage of a token
that role $r$ generates in $\tau_i$ is
\begin{equation}
\widehat A^{\mathrm{GRPO}}_{r,i}=\frac{R_{r,i}-\mu_{x,r}}{\sigma_{x,r}+\epsilon},
\label{eq:grpo}
\end{equation}
where $\mu_{x,r}$ and $\sigma_{x,r}$ are the mean and the standard deviation
of the rewards of role $r$ over the group of $x$ and $\epsilon$ is a small
constant. For the \answerer, the same normalization compares candidate
rewards within the parent workflow, yielding $\widehat A^{\mathrm{GRPO}}_{A,i,j}$.
Both stages maximize the PPO clipped surrogate \citep{schulman2017ppo}
using token advantages $\widehat A_t$, with different advantage assignments
in each stage.}

\subsection{Stage I: Learn a Teacher That Searches and Summarizes}
\label{sec:stage1}

\edit{In Stage~I, we train the simpler \searcher--\answerer workflow from $\pi_0$.
The \searcher receives the question, gathers evidence from a fixed
retriever and writes a report, and the \answerer answers the question from
the report. The \searcher thus helps the \answerer solve the question
directly, and the outcome of each trajectory depends on the evidence that one
\searcher gathers. Training the workflow teaches the model both searching
and summary writing.}

\edit{We reward the two roles with
\begin{equation}
\begin{aligned}
R_{A,i,j}&=c_{i,j},\\
R_{S,i}&=0.9\,R_{\mathrm{EM},i}+0.1\,b_{S,i},
\end{aligned}
\label{eq:stage1-reward}
\end{equation}
The behavior bonus $b_{S,i}$ equals 0.5 if the \searcher produces a summary
after one search step, 1 after two or more search steps, and 0 otherwise.
We assign a total reward of $-1$ to any output with an invalid action format. Answer correctness
thus carries most of the reward, and the \searcher earns a small
additional reward for its behavior. We compute the advantages of both roles
with Eq.~\eqref{eq:grpo}, update the model with PPO and
evaluate the trained policy in this cascaded workflow as C-MAS + RL.
We freeze the same Stage-I policy as the \teacher $\pi_T$ for Stage~II.}

\subsection{Stage II: Train the Team with Role-Specific Advantages}
\label{sec:stage2}

\begin{figure*}[!htbp]
\centering
\includegraphics[width=\textwidth]{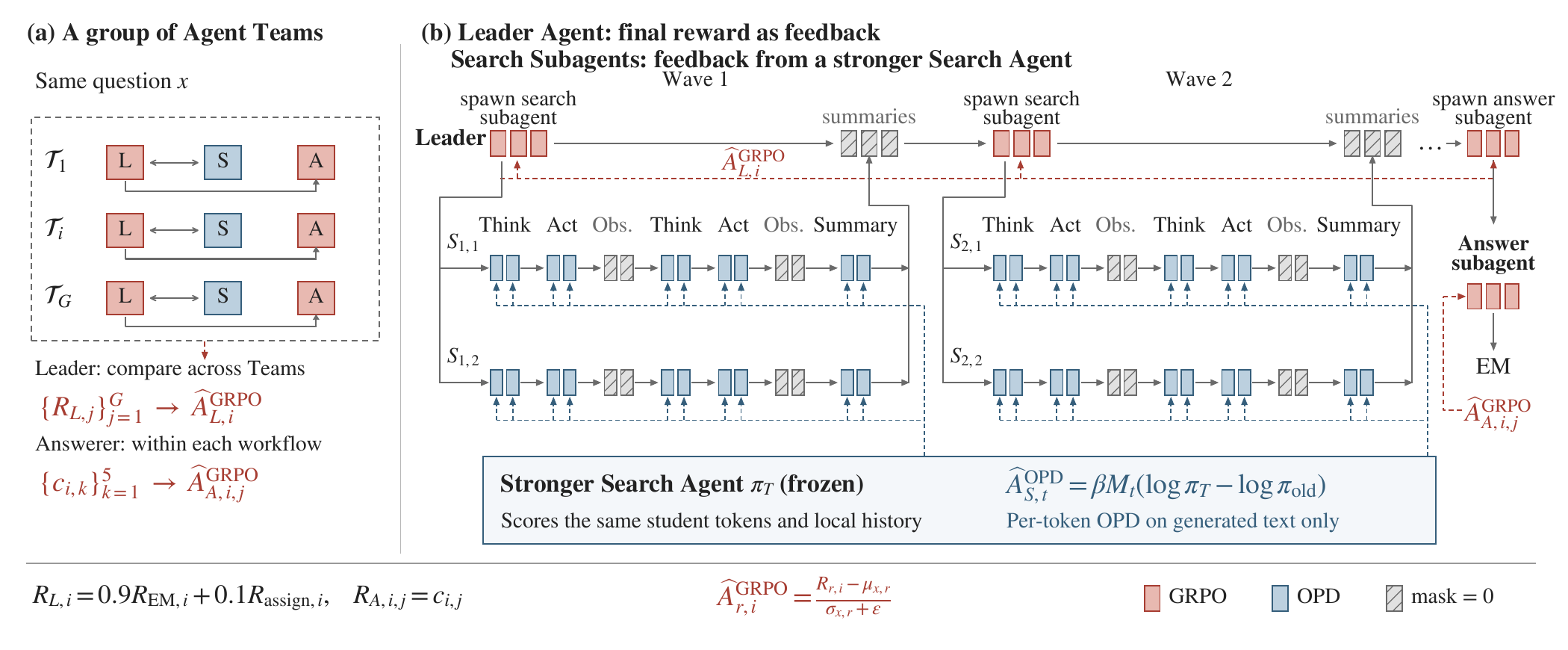}
\caption{\textbf{Role-selective process supervision in MultiAgent-OPD.}
(a) Multiple Agent Teams sample trajectories for the same question. Team outcomes yield GRPO advantages for the \leader, and candidate rewards yield GRPO advantages for the \answerer within each workflow.
(b) The \leader spawns parallel \searchers over successive waves, collects their summaries, and delegates the final answer to the \answerer. The \searchers follow ReAct trajectories to search and summarize results. The \searchers receive per-token OPD guidance weights from the frozen \teacher through teacher--old-student log-probability differences under the same local history.
Retrieved observations and summaries received by the \leader are masked from
the corresponding role's loss.}
\label{fig:credit-assignment}
\end{figure*}

\edit{In Stage~II, we initialize the hierarchical student from $\pi_0$ and
sample several team trajectories for each question, which form the group of
Eq.~\eqref{eq:grpo} (Figure~\ref{fig:credit-assignment}(a)). If the \searchers learned from the outcome reward, Eq.~\eqref{eq:grpo} would give all
\searchers of a trajectory the same advantage, whether or not the \leader
passes their reports to the \answerer. Eq.~\eqref{eq:grpo} would also
give every \searcher token zero advantage whenever the $G$ trajectories of a
question receive the same outcome. We train the \searchers with
token-level advantages from the \teacher. The \searcher role has the same
interface in both workflows, taking a question or a subtask and returning a
report, so the \teacher trained on questions can score the \searcher tokens of
the team on subtasks.}

\edit{We distill the \teacher into the \searchers with sampled reverse-KL
distillation \citep{agarwal2024gkd,gu2024minillm}. For each \searcher response,
we evaluate the frozen \teacher on the same role prompt, subtask, retrieved
passages and generated prefix as the student
(Figure~\ref{fig:credit-assignment}(b)). The distillation then minimizes
\begin{equation}
\begin{aligned}
\mathcal L_{\mathrm{OPD}}(\theta)&=\mathbb E_{\tau\sim\pi_\theta}\Big[\sum_t M_t\,\ell_t(\theta)\Big],\\
\ell_t(\theta)&=D_{\mathrm{KL}}\big(\pi_\theta(\cdot\mid h_t)\,\Vert\,\pi_T(\cdot\mid h_t)\big),
\end{aligned}
\label{eq:opd-obj}
\end{equation}
where the mask $M_t$ equals 1 on the tokens that the \searchers generate
and 0 elsewhere. The retriever writes the passages, so the mask excludes the
passages together with padding. At a fixed history, the negative gradient of
$\ell_t$ is
\begin{equation}
\begin{aligned}
-\nabla_\theta\ell_t(\theta)=\mathbb E_{a_t\sim\pi_\theta}\Big[&\log\frac{\pi_T(a_t\mid h_t)}{\pi_\theta(a_t\mid h_t)}\\
&\times\nabla_\theta\log\pi_\theta(a_t\mid h_t)\Big],
\end{aligned}
\label{eq:opd-gradient}
\end{equation}
since $\mathbb E_{a_t\sim\pi_\theta}[\nabla_\theta\log\pi_\theta(a_t\mid h_t)]=0$. The log-ratio thus acts as the advantage of the sampled token. In
the PPO update, each \searcher token $a_t$ sampled from
$\pi_{\mathrm{old}}$ receives the advantage
\begin{equation}
\widehat A^{\mathrm{OPD}}_{S,t}=\beta M_t\log\frac{\pi_T(a_t\mid h_t)}{\pi_{\mathrm{old}}(a_t\mid h_t)},
\label{eq:opd}
\end{equation}
where $\beta=0.01$.}

The log-ratio in Eq.~\eqref{eq:opd} serves as a teacher-guided local process-supervision weight. It reflects the frozen \teacher's preference at a student-generated history and does not estimate the token's causal contribution to the final answer. The advantage is positive when the \teacher assigns a token a higher probability than the sampling policy does, and the update then raises the probability of the token. Every \searcher token receives the signal, including tokens from trajectories whose final answer is wrong and tokens of questions on which all $G$ trajectories receive the same outcome. The \teacher scores prefixes that the student generates, so the supervision follows the new subtasks whenever the \leader changes the division of tasks.

\edit{The \leader and the \answerer keep Eq.~\eqref{eq:grpo} with outcome
rewards, since outcome rewards align the coordination of the \leader with team
success and evaluate the predictions of the \answerer directly. We reward
the \answerer with $R_{A,i,j}=c_{i,j}$ as in Stage~I and the
\leader with
\begin{equation}
R_{L,i}=0.9\,R_{\mathrm{EM},i}+0.1\,R_{\mathrm{assign},i},
\label{eq:leader-reward}
\end{equation}
where $R_{\mathrm{assign},i}=\min(C_{\mathrm{tool},i},6)/6$ and
$C_{\mathrm{tool},i}$ counts environment steps containing at least one
\searcher search action in $\tau_i$, so the \leader earns most of its reward
from correct answers and a bounded bonus for search activity. We normalize the rewards of the two roles
separately with Eq.~\eqref{eq:grpo} and give each token $a_t$ of a minibatch
the advantage of the role that generated the token,
\begin{equation}
\widehat A_t=
\begin{cases}
\widehat A^{\mathrm{GRPO}}_{L,i}, & r(a_t)=L,\\
\widehat A^{\mathrm{GRPO}}_{A,i,j}, & r(a_t)=A,\\
\widehat A^{\mathrm{OPD}}_{S,t}, & r(a_t)=S,
\end{cases}
\label{eq:role-adv}
\end{equation}
where $r(a_t)$ is the role that generated $a_t$ and $\tau_i$ is the
trajectory of $a_t$. The index $j$ identifies its \answerer candidate when $r(a_t)=A$.
We update the student with the PPO clipped objective
with the advantages of Eq.~\eqref{eq:role-adv} and an entropy bonus.
The frozen reference policy is the Stage-I \teacher $\pi_T$.
For the \leader and the \answerer, a KL penalty with coefficient
$10^{-3}$ toward $\pi_T$ enters the rewards before advantage computation.
For the \searchers, Eq.~\eqref{eq:opd} replaces the outcome-based
advantages with the teacher--student log-ratio. No additional KL penalty
is added to their policy loss. The \teacher serves only
during training, and at test time the student plays all three roles.}

\section{Experiments}
\label{sec:results}

We evaluate \method on open-domain question answering, training on NQ and HotpotQA and testing on 5,000 fixed questions from the seven benchmarks in Table~\ref{tab:main} \citep{kwiatkowski2019nq,joshi2017triviaqa,mallen2023popqa,yang2018hotpotqa,ho2020twoWiki,trivedi2022musique,press2023compositionality}. 

We compare \method with the cascaded and hierarchical baselines in the table and with five RL-based search agents, Search-R1 \citep{jin2025searchr1}, DeSA \citep{wang2025desa}, Tree-GRPO \citep{ji2026treegrpo}, MemSearcher \citep{yuan2025memsearcher} and Dr.\ MAS \citep{feng2026drmas}. We build all systems on Qwen3-4B-Instruct-2507 \citep{yang2025qwen3} and connect them to the same retriever, which returns three passages per query (Appendix~\ref{sec:impl}). The cascaded and hierarchical systems use inference limits of 5 and 12 environment steps, respectively (Appendix~\ref{sec:impl-settings}). 

We run each system four times per question, with one \answerer candidate per request, and score each final answer by normalized exact match. We report avg@4, the mean score over the four runs, and pass@4, the fraction of questions answered correctly at least once. avg@4 estimates the single-attempt success rate and requires no answer-selection rule, while pass@4 is the coverage upper bound across the four attempts. The primary metric is macro avg@4, which weights the seven datasets equally (Appendix~\ref{sec:appendix}).

\subsection{Main Results}
\label{sec:overall}

Table~\ref{tab:main} compares correctness across datasets for the cascaded and hierarchical systems and the five RL-based search agents.

\begin{table}[!htbp]
\reporttablesetup
\reportpanel{(a) avg@4 (\%)}
\begin{tabularx}{\linewidth}{>{\raggedright\arraybackslash\leavevmode}X*{8}{r}}
\reporttabletop
\rowcolor{TableGroup}
\reporthead{Method} & \multicolumn{3}{c}{\reporthead{Single-hop}} & \multicolumn{4}{c}{\reporthead{Multi-hop}} & \tablehead{Macro} \\
\rowcolor{TableGroup}
 & \reporthead{NQ} & \reporthead{TriviaQA} & \reporthead{PopQA} & \reporthead{HotpotQA} & \reporthead{2Wiki} & \reporthead{MuSiQue} & \reporthead{Bamboogle} & \\
\midrule
\multicolumn{9}{l}{\reportgroup{Cascaded and hierarchical systems}} \\[-1pt]
C-MAS base & 30.0 & 50.9 & 25.6 & 23.6 & 12.9 & 6.3 & 26.4 & 25.09 \\
C-MAS + RL & \textbf{43.7} & 60.2 & 33.9 & 33.3 & 30.5 & 11.5 & 39.4 & 36.07 \\
H-MAS base & 30.7 & 49.8 & 23.2 & 27.1 & 22.6 & 9.7 & 37.2 & 28.60 \\
Prompted Critic & 37.4 & 56.9 & 28.5 & 33.1 & 33.9 & 11.5 & 41.0 & 34.62 \\
\rowcolor{TableHighlight}
\textbf{MultiAgent-OPD} & 42.3 & \textbf{62.5} & 33.6 & \textbf{37.5} & 34.0 & \textbf{14.0} & \textbf{46.8} & \textbf{38.68} \\
\midrule
\multicolumn{9}{l}{\reportgroup{RL-based search agents}} \\[-1pt]
\edit{Search-R1} & \edit{41.1} & \edit{58.4} & \edit{32.8} & \edit{34.0} & \edit{32.1} & \edit{8.1} & \edit{27.2} & \edit{33.39} \\
\edit{DeSA} & \edit{35.9} & \edit{61.4} & \edit{35.9} & \edit{36.6} & \edit{33.0} & \edit{13.5} & \edit{42.4} & \edit{36.96} \\
\edit{Tree-GRPO} & \edit{38.6} & \edit{60.5} & \edit{36.3} & \edit{36.8} & \edit{32.8} & \edit{12.2} & \edit{44.4} & \edit{37.37} \\
\edit{MemSearcher} & \edit{37.8} & \edit{59.0} & \edit{\textbf{36.8}} & \edit{36.5} & \edit{\textbf{38.3}} & \edit{13.3} & \edit{36.0} & \edit{36.81} \\
\edit{Dr.\ MAS} & \edit{34.2} & \edit{60.5} & \edit{35.4} & \edit{36.5} & \edit{30.6} & \edit{13.2} & \edit{38.4} & \edit{35.54} \\
\bottomrule[0.6pt]
\end{tabularx}
\par\vspace{7pt}
\reportpanel{(b) pass@4 (\%)}
\begin{tabularx}{\linewidth}{>{\raggedright\arraybackslash\leavevmode}X*{8}{r}}
\reporttabletop
\rowcolor{TableGroup}
\reporthead{Method} & \multicolumn{3}{c}{\reporthead{Single-hop}} & \multicolumn{4}{c}{\reporthead{Multi-hop}} & \tablehead{Macro} \\
\rowcolor{TableGroup}
 & \reporthead{NQ} & \reporthead{TriviaQA} & \reporthead{PopQA} & \reporthead{HotpotQA} & \reporthead{2Wiki} & \reporthead{MuSiQue} & \reporthead{Bamboogle} & \\
\midrule
\multicolumn{9}{l}{\reportgroup{Cascaded and hierarchical systems}} \\[-1pt]
C-MAS base & 38.4 & 58.4 & 32.1 & 33.0 & 25.6 & 11.4 & 39.2 & 34.02 \\
C-MAS + RL & 53.8 & 67.6 & 40.9 & 44.0 & 46.7 & 19.1 & 49.6 & 45.97 \\
H-MAS base & 39.3 & 60.8 & 31.8 & 39.9 & 40.0 & 15.5 & 48.0 & 39.32 \\
Prompted Critic & 48.6 & 67.4 & 36.8 & 43.8 & 49.4 & 18.8 & 56.8 & 45.96 \\
\rowcolor{TableHighlight}
\textbf{MultiAgent-OPD} & 52.1 & \textbf{70.6} & 42.7 & 47.5 & 49.8 & \textbf{20.8} & \textbf{57.6} & \textbf{48.73} \\
\midrule
\multicolumn{9}{l}{\reportgroup{RL-based search agents}} \\[-1pt]
\edit{Search-R1} & \edit{50.5} & \edit{64.0} & \edit{39.9} & \edit{45.8} & \edit{49.2} & \edit{13.5} & \edit{43.2} & \edit{43.73} \\
\edit{DeSA} & \edit{51.5} & \edit{69.3} & \edit{42.0} & \edit{45.1} & \edit{47.2} & \edit{19.3} & \edit{55.2} & \edit{47.09} \\
\edit{Tree-GRPO} & \edit{\textbf{54.9}} & \edit{70.0} & \edit{40.5} & \edit{\textbf{48.7}} & \edit{48.3} & \edit{16.7} & \edit{53.6} & \edit{47.53} \\
\edit{MemSearcher} & \edit{53.0} & \edit{69.7} & \edit{\textbf{44.2}} & \edit{45.3} & \edit{\textbf{51.9}} & \edit{18.2} & \edit{51.2} & \edit{47.64} \\
\edit{Dr.\ MAS} & \edit{48.6} & \edit{68.1} & \edit{41.0} & \edit{43.7} & \edit{40.8} & \edit{15.1} & \edit{56.0} & \edit{44.76} \\
\bottomrule[0.6pt]
\end{tabularx}
\caption{\textbf{Main results: avg@4 and pass@4 (\%).} Macro weights the seven datasets equally. Missing and malformed answers score zero. C-MAS denotes the cascaded \searcher--\answerer system\edit{, and} H-MAS denotes the hierarchical \leader--\searcher--\answerer system. Prompted Critic and MultiAgent-OPD both use H-MAS. The two base systems use the same general-purpose instruction model without task-specific training. \edit{The RL-based search agents use the same backbone and protocol.} Bold marks the highest value for each metric and dataset.}
\label{tab:main}
\end{table}

\paragraph{The base hierarchy outperforms the base cascaded workflow.}
Both base systems directly use Qwen3-4B-Instruct-2507 without task-specific training. H-MAS, the hierarchical \leader--\searcher--\answerer system, reaches 28.60\% macro avg@4 versus C-MAS base's 25.09\%, a gain of 3.51 percentage points. Macro pass@4 rises from 34.02\% to 39.32\%. H-MAS has higher avg@4 on five of seven datasets, including all four multi-hop datasets, with the largest gains on Bamboogle and 2WikiMultiHopQA. 

\paragraph{C-MAS + RL has a higher score than Prompted Critic.}
C-MAS + RL, the Stage-I policy evaluated in its cascaded workflow, reaches 36.07\% macro avg@4. Prompted Critic, which trains the hierarchy with an M-GRPO-like combination of team rewards and a prompted judge of \searcher reports \citep{hong2025mgrpo}, reaches 34.62\%, 1.45 percentage points lower. Their macro pass@4 values are nearly tied (45.97\% and 45.96\%), so the observed difference is in average correctness across attempts. This comparison motivates evaluating local \searcher supervision within the hierarchy.

\paragraph{MultiAgent-OPD improves over both trained workflows.}
\method reaches 38.68\% macro avg@4, exceeding Prompted Critic by 4.06 percentage points and C-MAS + RL by 2.61 points. Both paired 95\% intervals exceed zero (Table~\ref{tab:paired-main}). The policy evaluated as C-MAS + RL supplies the frozen \teacher for \searcher OPD.

\begin{table}[!htbp]
\reporttablesetup
\begin{tabularx}{0.84\linewidth}{>{\raggedright\arraybackslash\leavevmode}Xrr}
\reporttabletop
\rowcolor{TableGroup}
\reporthead{Comparator} & \reporthead{$\Delta$ avg@4 (pp)} & \reporthead{95\% CI} \\
\midrule
Prompted Critic & +4.06 & [+2.98, +5.12] \\
C-MAS + RL & +2.61 & [+1.49, +3.73] \\
\bottomrule[0.6pt]
\end{tabularx}
\caption{MultiAgent-OPD score minus each compared system (percentage points). In each bootstrap sample, we resample questions within datasets and keep all four attempts for each question together.}
\label{tab:paired-main}
\end{table}

\label{sec:datasets}
\method exceeds Prompted Critic in avg@4 on all seven datasets and C-MAS + RL on five, with the largest observed gain over C-MAS + RL on Bamboogle (+7.40 percentage points). 

\paragraph{\edit{Comparison with RL-based search agents.}}
\method exceeds the strongest included agent, Tree-GRPO (37.37\%), by 1.31 percentage points in macro avg@4. On the single-hop and the multi-hop average alike, \method stays above each included agent, and the lead holds in macro pass@4 (48.73\% against at most 47.64\%).

\subsection{Diagnostic Comparisons}
\label{sec:ablation}

Table~\ref{tab:main-ablation} varies the \searcher learning signal, \teacher training, and direct supervision within the shared-parameter \leader--\searcher--\answerer hierarchy. All four configurations retain the \leader and \answerer objectives, normalization, and other training settings, including five answer candidates per request. Training batches differ; every row uses a step-500 checkpoint.

\begin{table}[!htbp]
\reporttablesetup
\reportpanel{(a) avg@4 (\%)}
\begin{tabularx}{\linewidth}{>{\raggedright\arraybackslash\leavevmode}X*{8}{r}}
\reporttabletop
\rowcolor{TableGroup}
\reporthead{Method} & \multicolumn{3}{c}{\reporthead{Single-hop}} & \multicolumn{4}{c}{\reporthead{Multi-hop}} & \tablehead{Macro} \\
\rowcolor{TableGroup}
 & \reporthead{NQ} & \reporthead{TriviaQA} & \reporthead{PopQA} & \reporthead{HotpotQA} & \reporthead{2Wiki} & \reporthead{MuSiQue} & \reporthead{Bamboogle} & \\
\midrule
\rowcolor{TableHighlight}
\textbf{MultiAgent-OPD} \textbf{(full)} & \textbf{42.3} & \textbf{62.5} & \textbf{33.6} & \textbf{37.5} & \textbf{34.0} & \textbf{14.0} & \textbf{46.8} & \textbf{38.68} \\
w/ outcome-based Searcher RL & 12.5 & 30.8 & 14.7 & 16.8 & 27.6 & 3.9 & 18.2 & 17.78 \\
w/ base-model teacher & 27.8 & 49.7 & 26.4 & 29.5 & 25.2 & 11.2 & 36.6 & 29.50 \\
w/o direct Searcher supervision & 31.3 & 51.7 & 27.1 & 28.2 & 21.2 & 10.1 & 39.2 & 29.82 \\
\bottomrule[0.6pt]
\end{tabularx}
\par\vspace{7pt}
\reportpanel{(b) pass@4 (\%)}
\begin{tabularx}{\linewidth}{>{\raggedright\arraybackslash\leavevmode}X*{8}{r}}
\reporttabletop
\rowcolor{TableGroup}
\reporthead{Method} & \multicolumn{3}{c}{\reporthead{Single-hop}} & \multicolumn{4}{c}{\reporthead{Multi-hop}} & \tablehead{Macro} \\
\rowcolor{TableGroup}
 & \reporthead{NQ} & \reporthead{TriviaQA} & \reporthead{PopQA} & \reporthead{HotpotQA} & \reporthead{2Wiki} & \reporthead{MuSiQue} & \reporthead{Bamboogle} & \\
\midrule
\rowcolor{TableHighlight}
\textbf{MultiAgent-OPD} \textbf{(full)} & \textbf{52.1} & \textbf{70.6} & \textbf{42.7} & \textbf{47.5} & \textbf{49.8} & \textbf{20.8} & \textbf{57.6} & \textbf{48.73} \\
w/ outcome-based Searcher RL & 17.0 & 38.3 & 17.9 & 20.3 & 34.2 & 6.3 & 25.6 & 22.79 \\
w/ base-model teacher & 37.7 & 62.3 & 35.1 & 40.7 & 39.6 & 18.6 & 52.0 & 40.86 \\
w/o direct Searcher supervision & 40.5 & 62.4 & 34.9 & 40.2 & 34.1 & 15.7 & 51.2 & 39.86 \\
\bottomrule[0.6pt]
\end{tabularx}
\caption{\textbf{Diagnostic comparisons of Searcher supervision and Teacher training: avg@4 and pass@4 (\%).} Relative to the full method, \textbf{w/ outcome-based Searcher RL} replaces \searcher OPD with RL using final-answer rewards. \textbf{w/ base-model teacher} uses the base model instead of the search-trained \teacher for OPD. \textbf{w/o direct Searcher supervision} removes the direct training losses of the \searchers. All variants keep the training of the \leader and the \answerer. Macro weights the seven datasets equally.}
\label{tab:main-ablation}
\end{table}

\textbf{w/ outcome-based Searcher RL} replaces \searcher OPD with group-relative advantages computed from final-answer exact-match rewards and removes distillation from the \teacher. \textbf{w/ base-model teacher} keeps \searcher OPD but uses an independently loaded, frozen base instruction model as the \teacher. \textbf{w/o direct Searcher supervision} sets \searcher policy, entropy, and direct KL losses to zero while keeping \searcher trajectories and the original loss denominator. Updates of the \leader and the \answerer still change the shared parameters used by the \searchers.

\paragraph{Outcome-based Searcher RL exhibits a collapse of search behavior.}
The outcome-based Searcher RL variant reaches 17.78\% macro avg@4, 10.82 points below H-MAS base. Across its 20,000 test workflows, no search action is recorded, although 99.51\% still produce subtasks and reports and 99.03\% produce a final answer. Development EM also declines as retrieval disappears (Appendix~\ref{sec:search-collapse}). This suggests that final-answer rewards failed to sustain retrieval in this hierarchical multi-agent system.

\paragraph{The task-trained-teacher configuration has higher scores.}
Replacing the OPD \teacher of Stage~II with the base model changes macro avg@4 from 38.68\% to 29.50\% and pass@4 from 48.73\% to 40.86\%. The full configuration scores higher on every dataset in both metrics. The base-model teacher variant keeps the token-level distillation objective, while replacing the task-trained \teacher that guides the student's queries and reports during hierarchical training.

\paragraph{The direct-supervision configuration has higher scores.}
Without the direct losses of the \searchers, macro avg@4 is 29.82\% and pass@4 is 39.86\%, compared with 38.68\% and 48.73\% for the full method. The \leader and the \answerer still receive their outcome rewards in this variant, and the \searchers still change through shared-parameter updates.

\subsection{Training Dynamics}
\label{sec:training-dynamics}

\begin{figure}[!htbp]
\centering
\includegraphics[width=0.72\textwidth,trim=0 6bp 0 8bp,clip]{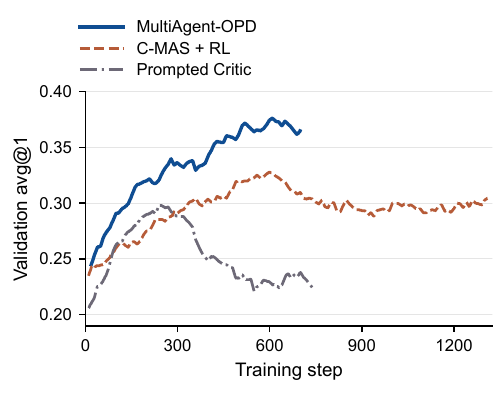}
\caption{\textbf{Training-time validation avg@1.} Each curve shows one smoothed historical validation run, digitized from the recorded curves. The horizontal axis counts training steps, not compute. These trajectories describe the displayed runs and do not measure variation across training seeds.}
\label{fig:training-dynamics}
\end{figure}

Figure~\ref{fig:training-dynamics} shows one historical validation trajectory for each method. The displayed \method trajectory rises through the middle of the run and remains above the two displayed baselines in their later shared interval. C-MAS + RL levels off, while Prompted Critic declines after an early increase.

\subsection{Answer Coverage and Consistency}
\label{sec:coverage}
\method reaches 48.73\% macro pass@4, compared with 45.97\% for C-MAS + RL and 45.96\% for \edit{Prompted Critic}. It also answers 29.27\% of questions correctly on all
four attempts, versus 27.05\% for C-MAS + RL and 23.24\% for \edit{Prompted Critic}
(Table~\ref{tab:coverage}). The gains \edit{thus} extend to both question
coverage and \edit{consistency}.

\begin{table}[!htbp]
\begingroup
\reporttablesetup
\begin{tabularx}{0.84\linewidth}{>{\raggedright\arraybackslash\leavevmode}Xrrr}
\reporttabletop
\rowcolor{TableGroup}
\reporthead{Method} & \reporthead{avg@4} & \reporthead{pass@4} & \reporthead{4/4 correct} \\
\midrule
C-MAS base & 25.09 & 34.02 & 17.12 \\
C-MAS + RL & 36.07 & 45.97 & 27.05 \\
H-MAS base & 28.60 & 39.32 & 18.73 \\
Prompted Critic & 34.62 & 45.96 & 23.24 \\
\rowcolor{TableHighlight}
\reportmethod & 38.68 & 48.73 & 29.27 \\
\bottomrule[0.6pt]
\end{tabularx}
\captionof{table}{\textbf{Success across four attempts (macro \%).} Pass@4 counts questions answered correctly at least once. The 4/4 score counts questions answered correctly every time. These scores do not include a method for selecting an answer.}
\label{tab:coverage}
\endgroup

\end{table}

\subsection{Inference Resource Usage}
\label{sec:cost}
\begin{figure}[!htbp]
\centering
\includegraphics[width=\textwidth,trim=0 5bp 0 4bp,clip]{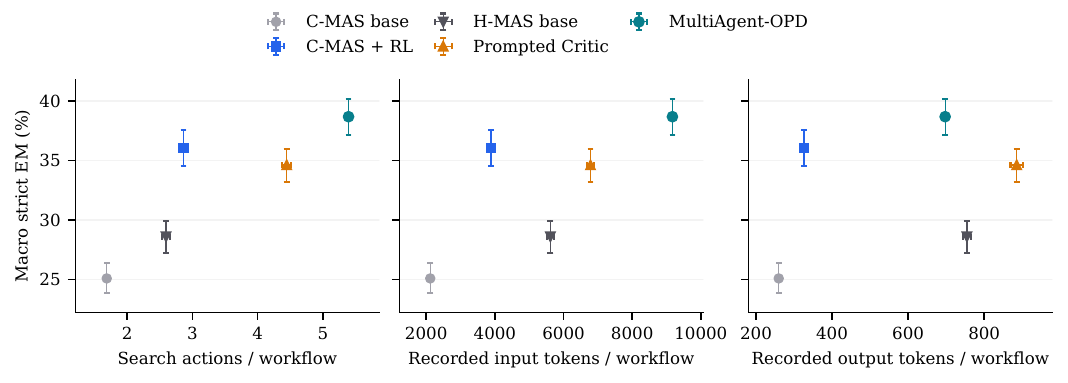}
\caption{Correctness and recorded inference workload. Recorded token totals include repeated histories and do not measure latency, unique context, service cost, or training compute. Error bars show pointwise question-bootstrap intervals.}
\label{fig:cost}
\end{figure}

The accuracy gains come with more search and input processing.
\method uses 5.39 search actions and 9,172 input tokens per workflow,
compared with 4.45 and 6,786 for \edit{Prompted Critic}, and 2.86 and 3,886 for
C-MAS + RL (Figure~\ref{fig:cost}\edit{,} dataset-macro averages). \edit{Relative to Prompted Critic, searches rise by 21.4\% and input tokens by 35.2\%, while output tokens fall from 885 to 697.} Recorded tokens include repeated histories \edit{(Appendix~\ref{sec:appendix})}. \edit{Prompted Critic and \method use the same \leader objective (Eq.~\eqref{eq:leader-reward}), so their workload difference reflects behavior learned under the same search-activity incentive. C-MAS + RL reaches 36.07\% macro avg@4 with the fewest searches, so a higher recorded workload alone does not explain the ranking.}

\FloatBarrier
\section{Conclusion}

\method jointly trains a hierarchical multi-agent team using a task-trained \teacher for local \searcher supervision, team success for the \leader, and candidate correctness for the \answerer. It achieves higher seven-dataset macro avg@4 and macro pass@4 than Prompted Critic, C-MAS + RL, and the included RL-based search agents, while using more searches and input tokens. A policy trained in a simpler workflow can thus guide subagents during joint training.

\section*{Limitations}

\paragraph{Scope and uncertainty.}
We evaluate cascaded and hierarchical systems and five RL-based search agents built on Qwen3-4B-Instruct-2507 on a fixed English open-domain QA set. Four attempts per question and question-bootstrap intervals characterize variation within the evaluated checkpoints, not variation across independently trained runs. Macro averaging weights the seven datasets equally, and normalized exact match does not detect semantic overlap or pretraining contamination. Systems run at their own workflow budgets. We do not compare them under a matched retrieval or token budget.

\paragraph{Mechanism.}
OPD supplies teacher-conditioned local process supervision, not a causal estimate of a token's contribution to the final answer. The teacher can make errors, full-question teacher training differs from delegated-subtask use, and shared role parameters couple updates across roles. Reverse-KL training can reduce exploration, and teacher and student require compatible token vocabularies. The diagnostic comparisons retain Leader and Answerer outcome objectives, so they do not separately evaluate those objectives.

\paragraph{Reproducibility.}
The anonymous artifact provides the scored evaluation records, aggregation script, result file, and evaluation protocol needed to reproduce the reported summary statistics. Recorded token counts measure inference workload.

\section*{Ethical Considerations}

We use public QA datasets and a text retriever. Retrieved material can be
inaccurate or biased. Exact match measures answer correctness. Further checks are needed to assess
whether evidence supports the answer, whether sources are credited, and
whether the system is safe to use.

\section*{AI Assistance Disclosure}

\edit{AI assistants} helped find related work, write and debug evaluation code, track
jobs, combine results, create figures, draft the paper, and review example
reports. The authors directed the research motivation and
argument of the revised manuscript. Quantitative results are computed from
recorded outputs. The human authors are responsible for the methods, numerical checks,
references, and final manuscript.

\FloatBarrier
\bibliographystyle{plainnat}
\bibliography{references}
\appendix
\section{Evaluation and Further Checks}
\label{sec:appendix}

\subsection{Fixed Test Set and Strict Scoring}

The sample was constructed with seed 20260728, before this evaluation.
NQ, TriviaQA, and PopQA each contribute 812 questions; HotpotQA,
2WikiMultiHopQA, and MuSiQue each contribute 813; Bamboogle contributes 125.
The split builder removed 14 test-pool entries matching normalized training
questions (10 TriviaQA, two NQ, two PopQA). Normalization applies Unicode NFKC,
lowercasing, punctuation removal, and whitespace collapse. No question hash is shared by the training, development, and main test sets. This check does not detect
paraphrases or pretraining contamination.

For each workflow, we use the last output from the \answerer. We take the text inside
its last \texttt{<answer>} tag and compare it with all accepted reference
answers. Before comparison, we convert text to lowercase and remove
punctuation, articles, and extra spaces. A missing span
scores zero. No workflow is discarded. Four workflows per question give
20,000 records per system and 100,000 across the five systems.

\subsection{Summary Scores and Uncertainty}

The anonymous artifact includes scores for each question and attempt, the aggregation script, and hashes of the included files. Readers can use it to check the summary results without a platform account. For each of 4,000 bootstrap draws (seed 20260912), questions are sampled with replacement within each dataset. A sampled question keeps its four attempts and all five methods. Dataset means receive equal weight. The intervals in Table~\ref{tab:ci} are the 2.5th and 97.5th percentiles; pairwise intervals are not adjusted for multiple comparisons. These intervals and the anonymous scored-record artifact cover C-MAS base, C-MAS + RL, H-MAS base, Prompted Critic, and \method. The five external RL-based search baselines are available as aggregate scores; comparisons with them are descriptive and have no paired intervals.

For the main comparison, checkpoints and the distillation coefficient were fixed before inspecting test results. The reported comparisons concern these retained policies.

To measure how often all four answers are correct, let $n_i=\sum_{k=1}^4c_{i,k}$,
where $c_{i,k}$ indicates whether attempt $k$ answers question $i$ correctly.
The 4/4-correct column is the dataset-macro average of
$N^{-1}\sum_i\mathbf{1}\{n_i=4\}$; the complete distribution of $n_i$ is
kept in the derived metrics. It reuses the existing four stochastic
workflows and introduces no additional sampling runs.

\begin{table}[!htbp]
\reporttablesetup
\begin{tabularx}{0.84\linewidth}{>{\raggedright\arraybackslash\leavevmode}Xrr}
\reporttabletop
\rowcolor{TableGroup}
\reporthead{Method} & \reporthead{avg@4 CI} & \reporthead{pass@4 CI} \\
\midrule
C-MAS base & [23.85, 26.37] & [32.41, 35.64] \\
C-MAS + RL & [34.54, 37.56] & [44.25, 47.65] \\
H-MAS base & [27.19, 29.95] & [37.63, 40.96] \\
Prompted Critic & [33.19, 35.99] & [44.25, 47.59] \\
\rowcolor{TableHighlight}
\reportmethod & [37.15, 40.18] & [47.05, 50.39] \\
\bottomrule[0.6pt]
\end{tabularx}
\caption{Macro scores with 95\% question-bootstrap intervals (\%) for the fixed models used in evaluation.}
\label{tab:ci}
\end{table}

\subsection{How We Count Resource Use}

Each workflow record contains prompt/response token totals, search actions,
returned passages, \searcher reports, \leader subtasks, and a duration value.
For each turn, the recorder applies the chat template to the assistant sample
and cuts it to the input and output limits. It then adds the token counts
across turns. This may count the same history more than once. Output counts
include EOS tokens. These counts let us compare recorded workload across
systems. They do not measure unique context, exact service usage, or total
training cost.

\searcher-action counts include nonempty queries recorded with a complete search
tag. We separately count search actions that return passages. A returned passage
may still be irrelevant or lack the needed evidence. The \searcher--\answerer workflow has no \leader, so the JSON value of zero for task assignments means this field does
not apply. It is not a measurement of \leader behavior. We also keep separate
report and task-assignment counts.

Costs average the four attempts within each question, then questions within
each dataset, then datasets equally. Relative changes are calculated from unrounded values. Pooled descriptive statistics are kept
separately. Batch duration is not a per-question response time measurement.

\subsection{Flexible Answer Extraction and the Completion Bound}

The extraction rules were selected before inspecting these diagnostic results. Table~\ref{tab:format} reports completion and flexible-extraction results from the saved
workflows, with the main scoring rule unchanged.

\begin{table}[!htbp]
\reporttablesetup
\begin{tabularx}{\linewidth}{>{\raggedright\arraybackslash\leavevmode}Xrrrrrr}
\reporttabletop
\rowcolor{TableGroup}
\reporthead{Method} & \shortstack{\reporthead{No}\\\reporthead{candidate}} & \shortstack{\reporthead{Empty}\\\reporthead{candidate}} & \reporthead{Untagged} & \shortstack{\reporthead{Failure}\\\reporthead{(macro \%)}} & \reporthead{Rescued} & \shortstack{\reporthead{Relaxed}\\\reporthead{EM (\%)}} \\
\midrule
C-MAS base & 493 & 0 & 173 & 3.41 & 3 & 25.105 \\
C-MAS + RL & 22 & 0 & 0 & 0.10 & 0 & 36.069 \\
H-MAS base & 2,451 & 104 & 109 & 12.79 & 5 & 28.624 \\
Prompted Critic & 658 & 9 & 0 & 3.73 & 0 & 34.617 \\
\rowcolor{TableHighlight}
\reportmethod & 62 & 66 & 0 & 0.68 & 0 & 38.677 \\
\bottomrule[0.6pt]
\end{tabularx}
\caption{\textbf{Answer completion and flexible extraction.} Candidate, untagged, and rescued are total counts out of 20,000 records per method. Failure rate and relaxed EM give each dataset equal weight. Rescued counts extra correct answers found by the flexible rules. All failures remain in the evaluation.}
\label{tab:format}
\end{table}

The flexible extractor applies the following rules in order. It does not use
the reference answer:
\begin{enumerate}
\item If the strict rule finds an answer, keep that answer unchanged.
\item Otherwise accept a paired answer tag with capitalization or internal
whitespace differences, using the last pair.
\item Otherwise accept a single unclosed answer opening tag only when followed
by 1--128 characters of single-line plain text.
\item Otherwise accept a trailing \texttt{Final answer:} or \texttt{Answer:}
line with a 1--128-character plain-text candidate.
\item Otherwise accept the entire output only if it is 1--128 characters,
single-line plain text without angle brackets or Markdown code fences.
\item If no rule applies, keep the record unparsed and incorrect.
\end{enumerate}
We score the extracted text using the original normalized EM rule and accepted
reference answers. These rules only change which formats we accept. They do
not judge meaning or use the reference to select an answer. The rescued count
is the total number of extra correct answers. Macro score differences give
each dataset equal weight.

We divide failures into three types: no \answerer candidate, an empty final
candidate, or a nonempty output without a valid answer tag. These types
describe the saved outputs, not the causes of failure. Each format failure
currently scores zero. If all such failures in the compared system became
correct, its score would be $a_B+f_B$. Subtracting this from \method's
unchanged score gives the bound in Table~\ref{tab:oracle}. We use the same
paired question samples for its intervals. This is an upper limit based on an
assumed repair of every failure, not a measured gain from actual repairs.

\begin{table}[!htbp]
\reporttablesetup
\begin{tabularx}{0.84\linewidth}{>{\raggedright\arraybackslash\leavevmode}Xrr}
\reporttabletop
\rowcolor{TableGroup}
\reporthead{Comparator} & \reporthead{Remaining gap (pp)} & \reporthead{95\% CI} \\
\midrule
Prompted Critic & +0.33 & [-0.80, +1.46] \\
C-MAS + RL & +2.51 & [+1.39, +3.65] \\
\bottomrule[0.6pt]
\end{tabularx}
\caption{\textbf{Upper bound from repairing format failures (pp).} We assume every format failure in the compared system becomes correct and leave MultiAgent-OPD unchanged. These are assumed bounds, not results from actual repairs.}
\label{tab:oracle}
\end{table}

\subsection{Collapse of Search Behavior under Outcome-Based Searcher RL}
\label{sec:search-collapse}

Table~\ref{tab:search-collapse} follows the outcome-based Searcher RL variant
on a fixed 1,000-question development set. Between steps 300 and 400, the
fraction of workflows with no recorded search rises from 0.5\% to 100\%,
while EM falls from 26.0\% to 21.8\%. Final-answer production increases to
99.3\%, indicating that the loss of retrieval occurs even as more workflows
finish with an answer. At step 500, retrieval remains absent and EM falls
to 19.2\%.

\begin{table}[!htbp]
\reporttablesetup
\begin{tabularx}{0.76\linewidth}{>{\raggedright\arraybackslash\leavevmode}Xrrr}
\reporttabletop
\rowcolor{TableGroup}
\reporthead{Step} & \reporthead{EM (\%)} & \reporthead{No search (\%)} & \reporthead{Final answer (\%)} \\
\midrule
100 & 27.0 & 1.2 & 64.8 \\
200 & 30.9 & 0.4 & 78.8 \\
300 & 26.0 & 0.5 & 49.9 \\
400 & 21.8 & 100.0 & 99.3 \\
500 & 19.2 & 100.0 & 99.5 \\
\bottomrule[0.6pt]
\end{tabularx}
\caption{Training progression of the outcome-based Searcher RL variant on
1,000 fixed development questions. EM, workflows with no recorded search,
and workflows producing a final answer are percentages. Development EM
uses one attempt per question and a pooled question average; it is distinct
from test-set macro avg@4.}
\label{tab:search-collapse}
\end{table}

The evaluated variant produces no recorded search actions in any of its
20,000 test workflows across the seven datasets. Nevertheless, 19,902
workflows (99.51\%) contain subtasks and reports, and 19,805 (99.03\%)
produce a final answer. These are pooled workflow percentages, rather than
dataset-macro averages. The recorder extracts search actions from
\texttt{<search>} tags independently of whether retrieval returns passages.
The observed failure is therefore a collapse of recorded search behavior
while report generation and answering continue. This supports the
interpretation that final-answer rewards did not sustain retrieval in this
configuration; it does not isolate the contribution of each training choice.

\Needspace{9\baselineskip}
\section{Systems and Training Details}
\label{sec:impl}

\subsection{Compared Systems}

The compared systems vary architecture, task-specific training, and \searcher supervision. C-MAS base and H-MAS base use the base model without task-specific training. C-MAS + RL jointly trains the \searcher and the \answerer with outcome-based RL in an implementation adapted from Dr.\ MAS \citep{feng2026drmas}; it evaluates the Stage-I policy in its original cascaded workflow. Prompted Critic trains the hierarchy with an M-GRPO-like combination of local and team rewards \citep{hong2025mgrpo}. The judge prompt appears in Appendix~\ref{sec:critic-prompt}. Its \searchers receive $0.3R_{\mathrm{EM}}+0.5R_{\mathrm{judge}}+0.2b_S$, where $R_{\mathrm{judge}}$ is a binary report score prompted from the shared checkpoint. Its \leader and \answerer use the same outcome objectives as in \method (Eq.~\eqref{eq:leader-reward}); the two trained hierarchies differ only in the \searcher learning signal. Published M-GRPO instead trains separate role models and aligns variable subagent trajectories. \method trains the same hierarchy with token-level \searcher OPD from the frozen Stage-I \teacher and outcome-based RL for the \leader and the \answerer. Neither trained hierarchy requires the \teacher or a judge at inference.

H-MAS base is not trained and therefore receives no Leader reward. The three diagnostic variants in Table~\ref{tab:main-ablation} retain the full configuration's Leader and Answerer objectives while changing the \searcher path.

\edit{The RL-based search agents use the same backbone and protocol as our
systems. Search-R1 trains a single agent with outcome rewards to interleave
reasoning and search calls. DeSA first trains search with a retrieval-recall
reward and then trains answering with an outcome reward. Tree-GRPO samples
trajectories as trees over agent steps and derives step-level advantages from
outcome rewards. MemSearcher keeps a compact memory across turns and trains
the agent with multi-context GRPO. Dr.\ MAS trains a hierarchical team with
agent-wise advantage normalization.}

\subsection{Training and Inference Settings}
\label{sec:impl-settings}

Training uses 79,168 Natural Questions and 90,447 HotpotQA examples
\citep{kwiatkowski2019nq,yang2018hotpotqa}. \edit{Training} uses batch size
32, five \edit{trajectories per question}, five PPO minibatch updates, and
learning rate $10^{-6}$, with veRL and SGLang
\citep{sheng2024hybridflow,zheng2024sglang}. Stage I allows five environment
steps; Stage II allows 12 steps and up to five \leader delegation rounds.
All training configurations, including the three diagnostic variants, use
five \answerer candidates per request. The full method and its ablations
share the same reward and advantage normalization and other training settings.
\answerer rewards are normalized within the parent workflow, as defined
in Eq.~\eqref{eq:grpo}. Evaluation uses one candidate per answering request
for all systems. The four evaluation runs per question are independent
workflows, each using this single-candidate answering rule.

Evaluation uses vLLM in float16, temperature 0.6, top-$p=0.95$, an 8,192-token input limit, and a 512-token output limit per call. Cascaded and hierarchical systems keep their respective 5- and 12-step limits. Following Dr.\ MAS \citep{feng2026drmas}, we use E5 \citep{wang2022e5} as the retriever. All systems use the same retriever, returning three passages per query. Saved trajectories provide search and token counts, completion statistics, and evidence-use examples.

\subsection{Evaluation Provenance and Statistics}
\label{sec:impl-provenance}

The main comparison uses a fixed 5,000-question manifest and four stochastic workflows per method. Appendix~\ref{sec:appendix} gives the dataset composition, question-overlap check, answer extraction rule, and question-bootstrap procedure. Checkpoints are saved every 100 steps. Trained systems in the main comparison use checkpoints nearest in step to the validation peak; all four rows of Table~\ref{tab:main-ablation} use step 500. Checkpoints and the OPD coefficient were fixed before their respective test evaluations. The anonymous artifact includes question--replica scores and a Python/NumPy script for recomputing macro metrics and paired intervals.

Figure~\ref{fig:training-dynamics} digitizes smoothed historical curves on a separate validation set, plotted against training steps rather than compute.

\section{Prompt Templates}
\label{sec:agent-prompts}
The following templates are used by the hierarchical team. For the \leader, \texttt{\{query\}} is the original question, and for a \searcher it is the delegated sub-question. The \answerer receives the original question and the reports selected by the \leader in \texttt{\{summaries\}}. C-MAS + RL uses the
same \searcher and \answerer templates, with the original question passed to the \searcher.
The role templates below reproduce the evaluation code; the critic template is used during Prompted Critic training.

\subsection{\leader}
\begin{promptcase}{casepurple}{Leader Agent Prompt}
\# Your Role\par You are a Leader Agent coordinating search and answer subagents to answer: \{query\}\par \smallskip\par \#\# Actions:\par - \textless{}search\_agent\textgreater{}...\textless{}/search\_agent\textgreater{} --- dispatch a search subagent. Put your specific sub-query inside the tags. Can send multiple at once.\par - \textless{}answer\_agent\textgreater{}...\textless{}/answer\_agent\textgreater{} --- produce final answer. Put relevant summary IDs (comma-separated) inside the tags.\par \smallskip\par \#\# Strategy:\par First briefly analyze what information is needed, then take action.\par \smallskip\par - Break multi-hop questions into single-fact sub-queries.\par - Write specific, targeted queries --- never vague or overly broad.\par - Never repeat queries already covered by existing summaries.\par \smallskip\par \#\# When to Act:\par - **Search** when: key entities unknown, relationships unclear, or facts unverified.\par - **Answer** when: summaries already contain all facts needed to fully answer the question.\par \smallskip\par \#\# Example Output:\par I need to find who won the 2024 Nobel Prize in Physics first.\par \textless{}search\_agent\textgreater{}2024 Nobel Prize in Physics winner\textless{}/search\_agent\textgreater{}\par \smallskip\par \#\# Example Output (multiple searches):\par I need both the director and the release year of this film.\par \textless{}search\_agent\textgreater{}Erta Ale documentary film director\textless{}/search\_agent\textgreater{}\par \textless{}search\_agent\textgreater{}Erta Ale documentary release year\textless{}/search\_agent\textgreater{}\par \smallskip\par \#\# Example Output (answer):\par The summaries contain all needed facts.\par \textless{}answer\_agent\textgreater{}01,03\textless{}/answer\_agent\textgreater{}
\end{promptcase}

\subsection{\searcher}
\begin{promptcase}{casebrown}{Search Agent Prompt}
\# Your Role\par You are a Search Agent. Iteratively search for information to answer: \{query\}\par \smallskip\par \#\# Actions (pick ONE per turn):\par - `\textless{}search\textgreater{}query\textless{}/search\textgreater{}` --- search for information. One query per turn. Fill a specific gap, avoid repeating past queries.\par - `\textless{}summary\textgreater{}...\textless{}/summary\textgreater{}` --- conclude with a structured summary when info is sufficient.\par \smallskip\par \#\# Rules:\par 1. Check past `\textless{}search\textgreater{}` and `\textless{}information\textgreater{}` to avoid redundant or too-broad queries.\par 2. When info is sufficient, output a `\textless{}summary\textgreater{}` with Key Findings and Relevance Assessment.
\end{promptcase}

\subsection{\answerer}
\begin{promptcase}{caseteal}{Answer Agent Prompt}
\# Your Role\par You are an Answer Agent. Use the summaries below and your own knowledge to answer the question.\par \smallskip\par \#\# Summaries\par \{summaries\}\par \smallskip\par \#\# Question\par \{query\}\par \smallskip\par \#\# Rules:\par 1. Then give a short, precise answer in `\textless{}answer\textgreater{}\textless{}/answer\textgreater{}`. Answer with the minimal necessary entity (name, date, number, phrase).
\end{promptcase}

\subsection{Prompted Critic}
\label{sec:critic-prompt}
The critic receives the original question in \texttt{\{question\}}, the delegated sub-question in \texttt{\{sub\_query\}}, and the search report in \texttt{\{summary\}}. The following template reproduces \texttt{CRITIC\_PROMPT} from the critic implementation.

\begin{promptcase}{casegray}{Critic Agent Prompt}
\# Your Role\par You are a Critic Agent. Your job is to evaluate a search agent's summary for a given sub-question derived from a main question.\par \smallskip\par \#\# Main Question\par \{question\}\par \smallskip\par \#\# Sub-Question Assigned to Search Agent\par \{sub\_query\}\par \smallskip\par \#\# Search Agent's Summary\par \{summary\}\par \smallskip\par \#\# Evaluation Criteria\par 1. **Relevance**: Does the summary directly address the sub-question?\par 2. **Completeness**: Does the summary cover the key aspects needed to answer the sub-question?\par 3. **Accuracy**: Is the information factually consistent and well-supported?\par 4. **Usefulness**: Will this summary help answer the main question?\par \smallskip\par \#\# Instructions\par 1. Output a binary score: `\textless{}score\textgreater{}1\textless{}/score\textgreater{}` if the summary is adequate, `\textless{}score\textgreater{}0\textless{}/score\textgreater{}` if it is not.\par 2. Output a concise, actionable hint in `\textless{}hint\textgreater{}...\textless{}/hint\textgreater{}` (10-30 words). If score=1, acknowledge what is good and note any minor gaps. If score=0, explain what is missing or wrong and suggest specific improvements.
\end{promptcase}

\section{Illustrative Examples of Multi-Agent Collaboration}
\label{sec:collaboration-cases}
We select two questions where H-MAS base and the C-MAS + RL
score 0/4 and \method scores 4/4. All examples show attempt 0 from the main
evaluation. Each search query is followed by all three stored document previews;
\texttt{[...]} marks text beyond the 200 characters saved by the recorder.
Search reports are verbatim excerpts. Messages are grouped by subagent;
this grouping does not specify the timing between subagents.

\Needspace{14\baselineskip}
\subsection{Resolving the Entity: Athlete and the River Thames}
\label{sec:case-1}
\Needspace{7\baselineskip}
\noindent\textbf{Case 1: H-MAS base}\par
\noindent{\small 0/4 correct; attempt 0.}\par
\label{casepage:1}
\smallskip
\begin{agentcase}{casegray}{0pt}{Query}
What is the body of water by the area where Athlete was formed?
\end{agentcase}
\begin{agentcase}{casepurple}{10pt}{Leader Agent}
01: What is the body of water near Athlete, Florida?\par 02: Is there a place named Athlete in Florida?
\end{agentcase}
\begin{agentcase}{casebrown}{10pt}{Search Agent 01}
what is the body of water near Athlete, Florida?
\end{agentcase}
\begin{agentcase}{casegray}{0pt}{Search Engine}
\textbf{Doc 1: 2010 NACAC Under-23 Championships in Athletics -- Results}\par 2010 NACAC Under-23 Championships in Athletics -- Results These are the results of the 2010 NACAC Under-23 Championships in At [...]\par \textbf{Doc 2: 2005 Central American and Caribbean Championships in Athletics -- Results}\par 10 July 10--11 2005 Central American and Caribbean Championships in Athletics -- Results These are the off [...]\par \textbf{Doc 3: 2010 NACAC Under-23 Championships in Athletics -- Results}\par -1.3 m/s, Heat 2: -1.1 m/s Final Wind: +2.2 m/s Heats Wind: Heat 1: +2.1 m/s, Heat 2: +3.5 m/s Final Wind: +2.1 m/s Final Final Fi [...]
\end{agentcase}
\begin{agentcase}{casebrown}{10pt}{Search Agent 01}
what is the nearest body of water to Athlete, Florida?
\end{agentcase}
\begin{agentcase}{casegray}{0pt}{Search Engine}
\textbf{Doc 1: Fort Lauderdale, Florida}\par Dolphins and the Miami Heat of the National Basketball Association all play in neighboring Dade County. The New York Yankees, Baltimore Orioles, and Kansas City Royals [...]\par \textbf{Doc 2: Florida Sports Hall of Fame}\par Florida Sports Hall of Fame The Florida Sports Hall of Fame (FSHOF) is an association dedicated to honoring athletes with outstanding achievement in sports in Flori [...]\par \textbf{Doc 3: Fort Lauderdale, Florida}\par one 20 by diving well. The complex is open to Fort Lauderdale residents, and has also been used in many different national and international competitions since its ope [...]
\end{agentcase}
\smallskip
\label{casepage:2}
\smallskip
\begin{agentcase}{casebrown}{10pt}{Search Agent 01}
geographical location of Athlete, Florida and nearby water bodies
\end{agentcase}
\begin{agentcase}{casegray}{0pt}{Search Engine}
\textbf{Doc 1: 2010 NACAC Under-23 Championships in Athletics -- Results}\par 2010 NACAC Under-23 Championships in Athletics -- Results These are the results of the 2010 NACAC Under-23 Championships in At [...]\par \textbf{Doc 2: 2010 NACAC Under-23 Championships in Athletics -- Results}\par -1.3 m/s, Heat 2: -1.1 m/s Final Wind: +2.2 m/s Heats Wind: Heat 1: +2.1 m/s, Heat 2: +3.5 m/s Final Wind: +2.1 m/s Final Final Fi [...]\par \textbf{Doc 3: 2005 Central American and Caribbean Championships in Athletics -- Results}\par 10 July 10--11 2005 Central American and Caribbean Championships in Athletics -- Results These are the off [...]
\end{agentcase}
\begin{agentcase}{casebrown}{10pt}{Search Agent 01}
location of Athlete, Florida and adjacent water bodies such as rivers, lakes, or oceans
\end{agentcase}
\noindent\begin{minipage}{\linewidth}
\begin{agentcase}[unbreakable]{casegray}{0pt}{Search Engine}
\textbf{Doc 1: Fort Lauderdale, Florida}\par Dolphins and the Miami Heat of the National Basketball Association all play in neighboring Dade County. The New York Yankees, Baltimore Orioles, and Kansas City Royals [...]\par \textbf{Doc 2: United States Olympic Training Center}\par United States Olympic Training Center The United States Olympic Training Centers (OTCs) are three campuses created by the United States Olympic Committee  [...]\par \textbf{Doc 3: United States Olympic Training Center}\par United States Olympic Committee will continue funding athlete programming at the center at least through 2020. United States Olympic Training Center The U [...]
\end{agentcase}
\end{minipage}
\begin{agentcase}{casebrown}{10pt}{Search Agent 01}
where is Athlete, Florida located and what is the nearest body of water?
\end{agentcase}
\begin{agentcase}{casegray}{0pt}{Search Engine}
\textbf{Doc 1: 2005 Central American and Caribbean Championships in Athletics -- Results}\par 10 July 10--11 2005 Central American and Caribbean Championships in Athletics -- Results These are the off [...]\par \textbf{Doc 2: 2010 NACAC Under-23 Championships in Athletics -- Results}\par 2010 NACAC Under-23 Championships in Athletics -- Results These are the results of the 2010 NACAC Under-23 Championships in At [...]\par \textbf{Doc 3: 2009 Central American and Caribbean Age Group Championships in Athletics}\par 2009 Central American and Caribbean Age Group Championships in Athletics The 13th Central American and Caribbean Age Gr [...]
\end{agentcase}
\smallskip
\label{casepage:3}
\smallskip
\begin{agentcase}{casebrown}{10pt}{Search Agent 02}
Is there a place named Athlete in Florida?
\end{agentcase}
\begin{agentcase}{casegray}{0pt}{Search Engine}
\textbf{Doc 1: Florida Sports Hall of Fame}\par Florida Sports Hall of Fame The Florida Sports Hall of Fame (FSHOF) is an association dedicated to honoring athletes with outstanding achievement in sports in Flori [...]\par \textbf{Doc 2: Coastal Florida Sports Park}\par with a great deal of support from private sports investors, the Cocoa Expo Sports Center received a long-overdue renovation of the entire complex. From the stadium  [...]\par \textbf{Doc 3: Florida Gators track and field}\par The following Florida Gators athletes have earned medals in one or more track and field events at the Summer Olympics: Florida Gators track and field The Florida [...]
\end{agentcase}
\begin{agentcase}{casebrown}{10pt}{Search Agent -- Report}
There is no geographical location or town named "Athlete" in Florida.\par The term "Athlete" appears in contexts such as the "Florida Sports Hall of Fame," which honors accomplished athletes from Florida, and in references to athletic programs like the Florida Gators track and field team.
\end{agentcase}
\begin{agentcase}{casepurple}{10pt}{Leader Agent}
Selected reports: 02
\end{agentcase}
\noindent\begin{minipage}{\linewidth}
\begin{agentcase}[unbreakable]{caseteal}{10pt}{Answer Agent}
\textless{}answer\textgreater{}There is no place named Athlete in Florida, so there is no body of water associated with its formation.\textless{}/answer\textgreater{}
\end{agentcase}
\trajectoryend
\end{minipage}
\Needspace{7\baselineskip}
\noindent\textbf{Case 1: C-MAS + RL}\par
\noindent{\small 0/4 correct; attempt 0.}\par
\label{casepage:4}
\smallskip
\begin{agentcase}{casegray}{0pt}{Query}
What is the body of water by the area where Athlete was formed?
\end{agentcase}
\begin{agentcase}{casebrown}{10pt}{Search Agent 01}
where is Athlete formed?
\end{agentcase}
\begin{agentcase}{casegray}{0pt}{Search Engine}
\textbf{Doc 1: Athlete}\par definition of ""sportsman"" according to Webster's ""Third Unabridged Dictionary"" (1960) is, ""a person who is active in sports: as (a): one who engages in the sports of the [...]\par \textbf{Doc 2: Athletic heart syndrome}\par Athletic heart syndrome Athletic heart syndrome (AHS), also known as athlete's heart, athletic bradycardia, or exercise-induced cardiomegaly is a non-pathological condi [...]\par \textbf{Doc 3: Health issues in athletics}\par Health issues in athletics Health issues of athletics concern the health and well-being of athletes who participate in an organized sport. If athletes are physically [...]
\end{agentcase}
\begin{agentcase}{casebrown}{10pt}{Search Agent 01}
Body of water near the town or location named Athlete?**
\end{agentcase}
\begin{agentcase}{casegray}{0pt}{Search Engine}
\textbf{Doc 1: Team sport}\par match surrounded by crowds. Prehistoric cave-paintings in Japan show a sport similar to sumo wrestling. In Wadi Sura, near Gilf Kebir in Libya, a Neolithic rock painting in the cave [...]\par \textbf{Doc 2: International Olympic Academy}\par for so many years in effect proved that the correct and effective means for the special Olympic education of the people who would wish to offer their strength and [...]\par \textbf{Doc 3: NC Vouliagmeni}\par NC Vouliagmeni The Nautical Club of Vouliagmeni, or ""NOV"" (Greek: {\fontfamily{cmr}\selectfont\foreignlanguage{greek}{Ναυτικός}} {\fontfamily{cmr}\selectfont\foreignlanguage{greek}{Όμιλος}} {\fontfamily{cmr}\selectfont\foreignlanguage{greek}{Βου}}\textbackslash{} [...]
\end{agentcase}
\smallskip
\label{casepage:5}
\smallskip
\begin{agentcase}{casebrown}{10pt}{Search Agent 01}
Near which body of water is the place named Athlete located? Or is there a geographical location called Athlete?
\end{agentcase}
\begin{agentcase}{casegray}{0pt}{Search Engine}
\textbf{Doc 1: Athlete}\par Athlete An athlete (also sportsman or sportswoman) is a person who competes in one or more sports that involve physical strength, speed or endurance. The application of the term to tho [...]\par \textbf{Doc 2: Athlete}\par definition of ""sportsman"" according to Webster's ""Third Unabridged Dictionary"" (1960) is, ""a person who is active in sports: as (a): one who engages in the sports of the [...]\par \textbf{Doc 3: Athlete}\par soccer, baseball, lacrosse, and did ballroom dancing. King Gustav V of Sweden, while awarding Thorpe the decathlon gold said: ""You, sir, are the greatest athlete in the world."" T [...]
\end{agentcase}
\begin{agentcase}{casebrown}{10pt}{Search Agent -- Report}
The term "athlete" refers to a person engaged in physical sports or competitive activities.\par There is no geographical location or body of water named "Athlete" based on available information.
\end{agentcase}
\noindent\begin{minipage}{\linewidth}
\begin{agentcase}[unbreakable]{caseteal}{10pt}{Answer Agent}
\textless{}answer\textgreater{}Sea of Greece\textless{}/answer\textgreater{}
\end{agentcase}
\trajectoryend
\end{minipage}
\Needspace{7\baselineskip}
\noindent\textbf{Case 1: MultiAgent-OPD}\par
\noindent{\small 4/4 correct; attempt 0.}\par
\label{casepage:6}
\smallskip
\begin{agentcase}{casegray}{0pt}{Query}
What is the body of water by the area where Athlete was formed?
\end{agentcase}
\begin{agentcase}{casepurple}{10pt}{Leader Agent}
01: who formed the band Athlete\par 02: what is the body of water near Deptford, London
\end{agentcase}
\begin{agentcase}{casebrown}{10pt}{Search Agent 01}
who formed the band Athlete
\end{agentcase}
\begin{agentcase}{casegray}{0pt}{Search Engine}
\textbf{Doc 1: Wilt (band)}\par Wilt (band) Wilt were an alternative rock band founded in Kilkenny, Ireland in 1998. Formed by ex-Kerbdog members Cormac Battle on vocals/guitar and Darragh Butler on drums, the li [...]\par \textbf{Doc 2: Wesley Geer}\par and groups, playing guitar, co-writing and/or producing songs, such as Papa Roach, The Damages, Ben Grosse, and Thrill Kill Kult. Born in Fullerton, California to parents Wesley Ha [...]\par \textbf{Doc 3: Matthew Meschery}\par Matthew Meschery Matthew Thomas Meschery Shakey Lo the Creation Kid (born April 12, 1975) is a vocalist and programmer. He is best known for being the original lead vocalist o [...]
\end{agentcase}
\begin{agentcase}{casebrown}{10pt}{Search Agent 01}
who formed the band Athlete (music group)
\end{agentcase}
\begin{agentcase}{casegray}{0pt}{Search Engine}
\textbf{Doc 1: Athlete (band)}\par Athlete (band) Athlete are an English indie rock band formed in Deptford, London, comprising Joel Pott (lead vocals and guitar), Carey Willetts (bass and backing vocals), Stephe [...]\par \textbf{Doc 2: Athlete (band)}\par records on regular basis. Their subsequent two albums ""Beyond the Neighbourhood"" (2007) and ""Black Swan"" (2009) sees the band exploring different styles, influenced  [...]\par \textbf{Doc 3: Sportsman (artist)}\par Sportsman (artist) Sportsman, real name Per Magnusson, is an artist, songwriter and previous singer in the Swedish band The Sonnets. He has released one single and one EP on [...]
\end{agentcase}
\begin{agentcase}{casebrown}{10pt}{Search Agent -- Report 01}
The band Athlete is an English indie rock band formed in Deptford, London.
\end{agentcase}
\smallskip
\label{casepage:7}
\smallskip
\begin{agentcase}{casebrown}{10pt}{Search Agent 02}
body of water near Deptford, London
\end{agentcase}
\begin{agentcase}{casegray}{0pt}{Search Engine}
\textbf{Doc 1: Deptford}\par Deptford Deptford ( ) is a district of south-east London, England, within the London Borough of Lewisham. From the mid-16th to the late 19th century, Deptford was home to Deptford Doc [...]\par \textbf{Doc 2: Deptford Wharf}\par Deptford Wharf Deptford Wharf in London, UK is situated on the Thames Path south-east of South Dock Marina, across the culverted mouth of the Earl's Sluice and north of Aragon T [...]\par \textbf{Doc 3: Deptford}\par Evelyn in the north and part of New Cross to the south. Deptford borders the areas of Brockley and Lewisham to the south, New Cross to the west and Rotherhithe to the north west; Dept [...]
\end{agentcase}
\begin{agentcase}{casebrown}{10pt}{Search Agent 02}
River Thames
\end{agentcase}
\begin{agentcase}{casegray}{0pt}{Search Engine}
\textbf{Doc 1: River Thames}\par River Thames The River Thames, ( ) known alternatively in parts as the Isis, is a river that flows through southern England including London. At , it is the longest river entirely [...]\par \textbf{Doc 2: River Thames}\par river a number of smaller private companies also offer river trips at Oxford, Wallingford, Reading and Hampton Court. Many companies also provide boat hire on the river. The leisu [...]\par \textbf{Doc 3: River Thames}\par to help reduce the flood risk in Maidenhead Windsor and Eton. In 2010, the Thames won the largest environmental award in the world -- the \$350,000 International Riverprize. On [...]
\end{agentcase}
\begin{agentcase}{casebrown}{10pt}{Search Agent -- Report 02}
The primary body of water near Deptford, London is the River Thames.\par Deptford is situated on the south bank of the River Thames, specifically in the area known as North Deptford, which is the only part of the London Borough of Lewisham to front the Thames.
\end{agentcase}
\begin{agentcase}{casepurple}{10pt}{Leader Agent}
Selected reports: 01,02
\end{agentcase}
\noindent\begin{minipage}{\linewidth}
\begin{agentcase}[unbreakable]{caseteal}{10pt}{Answer Agent}
\textless{}answer\textgreater{}River Thames\textless{}/answer\textgreater{}
\end{agentcase}
\trajectoryend
\end{minipage}
\par\smallskip{\fontsize{9.3}{10.6}\selectfont H-MAS base searches for a place called Athlete in Florida. The C-MAS + RL system also treats Athlete as a place or a generic sports term. MultiAgent-OPD identifies Athlete as a band, reports its origin in Deptford, and connects Deptford to the River Thames. Its two search subtasks provide the facts needed for the final answer.\par}

\Needspace{10\baselineskip}
\subsection{Preserving the Caretaker-Role Constraint}
\label{sec:case-2}
\Needspace{7\baselineskip}
\noindent\textbf{Case 2: H-MAS base}\par
\noindent{\small 0/4 correct; attempt 0.}\par
\label{casepage:8}
\smallskip
\begin{agentcase}{casegray}{0pt}{Query}
Who succeeded Alf Ramsey as England (caretaker) football manager in 1974?
\end{agentcase}
\begin{agentcase}{casepurple}{10pt}{Leader Agent}
01: Who succeeded Alf Ramsey as England caretaker football manager in 1974?
\end{agentcase}
\begin{agentcase}{casebrown}{10pt}{Search Agent 01}
Who succeeded Alf Ramsey as England caretaker football manager in 1974?
\end{agentcase}
\begin{agentcase}{casegray}{0pt}{Search Engine}
\textbf{Doc 1: Alf Ramsey}\par Alf Ramsey Sir Alfred Ernest Ramsey (22 January 1920 -- 28 April 1999) was an English football player and manager who was manager of England from 1963 to 1974, and guided Englan [...]\par \textbf{Doc 2: Alf Ramsey}\par FA). England, having won the World Cup in 1966, were now perceived to have failed in three successive tournaments. The disappointments of quarter final exits from major tournaments  [...]\par \textbf{Doc 3: 1973--74 Football League}\par 1973--74 Football League The 1973--74 season was the 75th completed season of The Football League. Don Revie marked his last season as Leeds United's manag [...]
\end{agentcase}
\begin{agentcase}{casebrown}{10pt}{Search Agent -- Report}
Don Revie succeeded Ramsey as England's national football team manager, taking over immediately after Ramsey's departure.\par The information confirms that Don Revie succeeded Alf Ramsey as England's caretaker (and subsequent) manager in 1974, directly addressing the query. The documents provide clear historical context and confirm the succession without ambiguity.
\end{agentcase}
\begin{agentcase}{casepurple}{10pt}{Leader Agent}
Selected reports: 01
\end{agentcase}
\noindent\begin{minipage}{\linewidth}
\begin{agentcase}[unbreakable]{caseteal}{10pt}{Answer Agent}
\textless{}answer\textgreater{}Don Revie\textless{}/answer\textgreater{}
\end{agentcase}
\trajectoryend
\end{minipage}
\Needspace{7\baselineskip}
\noindent\textbf{Case 2: C-MAS + RL}\par
\noindent{\small 0/4 correct; attempt 0.}\par
\label{casepage:9}
\smallskip
\begin{agentcase}{casegray}{0pt}{Query}
Who succeeded Alf Ramsey as England (caretaker) football manager in 1974?
\end{agentcase}
\begin{agentcase}{casebrown}{10pt}{Search Agent 01}
Who succeeded Alf Ramsey as England caretaker manager in 1974?
\end{agentcase}
\begin{agentcase}{casegray}{0pt}{Search Engine}
\textbf{Doc 1: Alf Ramsey}\par Alf Ramsey Sir Alfred Ernest Ramsey (22 January 1920 -- 28 April 1999) was an English football player and manager who was manager of England from 1963 to 1974, and guided Englan [...]\par \textbf{Doc 2: Alf Ramsey}\par FA). England, having won the World Cup in 1966, were now perceived to have failed in three successive tournaments. The disappointments of quarter final exits from major tournaments  [...]\par \textbf{Doc 3: Alf Ramsey}\par his England side being dubbed ""The Wingless Wonders"". England's World Cup victory at Wembley in 1966 made Ramsey a national hero, though he had his critics, both at the time a [...]
\end{agentcase}
\begin{agentcase}{casebrown}{10pt}{Search Agent 01}
Who took over as caretaker manager of England after Alf Ramsey was sacked in 1974?
\end{agentcase}
\begin{agentcase}{casegray}{0pt}{Search Engine}
\textbf{Doc 1: Alf Ramsey}\par FA). England, having won the World Cup in 1966, were now perceived to have failed in three successive tournaments. The disappointments of quarter final exits from major tournaments  [...]\par \textbf{Doc 2: Alf Ramsey}\par Alf Ramsey Sir Alfred Ernest Ramsey (22 January 1920 -- 28 April 1999) was an English football player and manager who was manager of England from 1963 to 1974, and guided Englan [...]\par \textbf{Doc 3: 1973--74 in English football}\par 1973--74 in English football The 1973--74 season was the 94th season of competitive football in England. It is considered as the end of Leeds United do [...]
\end{agentcase}
\smallskip
\label{casepage:10}
\smallskip
\begin{agentcase}{casebrown}{10pt}{Search Agent 01}
Geographic distribution of ozone layers
\end{agentcase}
\begin{agentcase}{casegray}{0pt}{Search Engine}
\textbf{Doc 1: Ozone layer}\par Ozone layer The Ozone layer or ozone shield is a region of Earth's stratosphere that absorbs most of the Sun's ultraviolet radiation. It contains high concentrations of ozone (O) i [...]\par \textbf{Doc 2: Ozone layer}\par 28, 2003, which operated until early 2012. Ozone layer The Ozone layer or ozone shield is a region of Earth's stratosphere that absorbs most of the Sun's ultraviolet radiation. It  [...]\par \textbf{Doc 3: Ozone layer}\par called the ozone-oxygen cycle. Chemically, this can be described as: About 90 percent of the ozone in the atmosphere is contained in the stratosphere. Ozone concentrations are grea [...]
\end{agentcase}
\begin{agentcase}{casebrown}{10pt}{Search Agent -- Report}
There is a mention that Don Revie, who managed Leeds United, took over from Sir Alf Ramsey as England national team manager after Ramsey's sacking in 1974.\par The information confirms that Don Revie succeeded Alf Ramsey as caretaker manager of England in 1974. However, cross-referencing the details: While Revie is noted to have managed Leeds United during the 1973--74 season and succeeded Ramsey in a managerial capacity, the term "caretaker manager" is not explicitly used in the documents. The transition from Ramsey to Revie is presented in the context of a managerial handover, but without explicit confirmation of the "caretaker" designation in official sources or FA records. Therefore, while Don Revie is the most plausible successor based on available information, the specific label "caretaker manager" may not be historically accurate in this context. A more precise source on the official FA designation would be required for absolute certainty.
\end{agentcase}
\noindent\begin{minipage}{\linewidth}
\begin{agentcase}[unbreakable]{caseteal}{10pt}{Answer Agent}
\textless{}answer\textgreater{}Don Revie\textless{}/answer\textgreater{}
\end{agentcase}
\trajectoryend
\end{minipage}
\Needspace{7\baselineskip}
\noindent\textbf{Case 2: MultiAgent-OPD}\par
\noindent{\small 4/4 correct; attempt 0.}\par
\label{casepage:11}
\smallskip
\begin{agentcase}{casegray}{0pt}{Query}
Who succeeded Alf Ramsey as England (caretaker) football manager in 1974?
\end{agentcase}
\begin{agentcase}{casepurple}{10pt}{Leader Agent}
01: who succeeded Alf Ramsey as caretaker manager of England national football team in 1974\par 02: what is the duration of Joe Mercer's caretaker management of England
\end{agentcase}
\begin{agentcase}{casebrown}{10pt}{Search Agent 01}
who succeeded Alf Ramsey as caretaker manager of England national football team in 1974
\end{agentcase}
\begin{agentcase}{casegray}{0pt}{Search Engine}
\textbf{Doc 1: Alf Ramsey}\par Alf Ramsey Sir Alfred Ernest Ramsey (22 January 1920 -- 28 April 1999) was an English football player and manager who was manager of England from 1963 to 1974, and guided Englan [...]\par \textbf{Doc 2: Alf Ramsey}\par managers ever in ""The Independent"". He is widely regarded as one of British football's all-time great managers. Nonetheless, not everyone reveres Ramsey's managerial style. Ac [...]\par \textbf{Doc 3: History of the England national football team}\par equalised from the penalty spot six minutes later and chances continued to rain down on the Polish goal, the score remained 1--1. Although at  [...]
\end{agentcase}
\begin{agentcase}{casebrown}{10pt}{Search Agent 01}
who served as caretaker manager of England after Alf Ramsey was sacked in 1974
\end{agentcase}
\begin{agentcase}{casegray}{0pt}{Search Engine}
\textbf{Doc 1: Alf Ramsey}\par Alf Ramsey Sir Alfred Ernest Ramsey (22 January 1920 -- 28 April 1999) was an English football player and manager who was manager of England from 1963 to 1974, and guided Englan [...]\par \textbf{Doc 2: Alf Ramsey}\par FA). England, having won the World Cup in 1966, were now perceived to have failed in three successive tournaments. The disappointments of quarter final exits from major tournaments  [...]\par \textbf{Doc 3: Alf Ramsey}\par managers ever in ""The Independent"". He is widely regarded as one of British football's all-time great managers. Nonetheless, not everyone reveres Ramsey's managerial style. Ac [...]
\end{agentcase}
\smallskip
\label{casepage:12}
\smallskip
\begin{agentcase}{casebrown}{10pt}{Search Agent 01}
who was the caretaker manager of England after Alf Ramsey's sacking in 1974
\end{agentcase}
\begin{agentcase}{casegray}{0pt}{Search Engine}
\textbf{Doc 1: Alf Ramsey}\par Alf Ramsey Sir Alfred Ernest Ramsey (22 January 1920 -- 28 April 1999) was an English football player and manager who was manager of England from 1963 to 1974, and guided Englan [...]\par \textbf{Doc 2: Alf Ramsey}\par FA). England, having won the World Cup in 1966, were now perceived to have failed in three successive tournaments. The disappointments of quarter final exits from major tournaments  [...]\par \textbf{Doc 3: Alf Ramsey}\par managers ever in ""The Independent"". He is widely regarded as one of British football's all-time great managers. Nonetheless, not everyone reveres Ramsey's managerial style. Ac [...]
\end{agentcase}
\begin{agentcase}{casebrown}{10pt}{Search Agent -- Report 01}
After Ramsey's sacking, Joe Mercer served as caretaker manager of England.\par Don Revie was appointed as Ramsey's permanent successor by the FA.
\end{agentcase}
\begin{agentcase}{casebrown}{10pt}{Search Agent 02}
Joe Mercer's caretaker management duration of England national team
\end{agentcase}
\begin{agentcase}{casegray}{0pt}{Search Engine}
\textbf{Doc 1: England national football team manager}\par ended in a 1--1 draw. England had failed to qualify for the World Cup, and Ramsey was dismissed the following May. Joe Mercer took control of the tea [...]\par \textbf{Doc 2: England national football team manager}\par on third party ownership of players. He subsequently left the role as manager of England later that day. Former England international defender and Englan [...]\par \textbf{Doc 3: England national football team manager}\par inception; four of those were in short-term caretaker manager roles: Joe Mercer (seven games in charge), Howard Wilkinson (two games, a year apart from o [...]
\end{agentcase}
\smallskip
\label{casepage:13}
\smallskip
\begin{agentcase}{casebrown}{10pt}{Search Agent 02}
Joe Mercer's caretaker management duration of England national team (exact number of games or days)
\end{agentcase}
\begin{agentcase}{casegray}{0pt}{Search Engine}
\textbf{Doc 1: England national football team manager}\par ended in a 1--1 draw. England had failed to qualify for the World Cup, and Ramsey was dismissed the following May. Joe Mercer took control of the tea [...]\par \textbf{Doc 2: England national football team manager}\par inception; four of those were in short-term caretaker manager roles: Joe Mercer (seven games in charge), Howard Wilkinson (two games, a year apart from o [...]\par \textbf{Doc 3: Steve McClaren}\par first time in 14 years that England had not qualified for a major tournament, and the first time in 24 years that they had not qualified for the European Championships. The foll [...]
\end{agentcase}
\begin{agentcase}{casebrown}{10pt}{Search Agent -- Report 02}
Joe Mercer served as caretaker manager of the England national football team for seven matches.\par His tenure followed the dismissal of Alf Ramsey and preceded the appointment of Don Revie.
\end{agentcase}
\begin{agentcase}{casepurple}{10pt}{Leader Agent}
Selected reports: 01,02
\end{agentcase}
\noindent\begin{minipage}{\linewidth}
\begin{agentcase}[unbreakable]{caseteal}{10pt}{Answer Agent}
\textless{}answer\textgreater{}Joe Mercer\textless{}/answer\textgreater{}
\end{agentcase}
\trajectoryend
\end{minipage}
\par\smallskip{\fontsize{9.3}{10.6}\selectfont Both comparison systems answer Don Revie. Their reports confuse or leave unresolved the distinction between a caretaker and a permanent manager. MultiAgent-OPD separates Joe Mercer's caretaker role from Don Revie's permanent appointment, and its second subtask checks Mercer's seven-match tenure.\par}

\end{document}